\documentclass[10pt]{siamltex}

\usepackage{ifthen,algorithm,algorithmic}
\usepackage{amssymb,amsfonts,amsmath,latexsym,dsfont}
\usepackage{tikz,pgfplots,pgflibraryplotmarks}
\usetikzlibrary{calc}
\usepackage{fullpage}
\usepackage{graphicx}
\usepackage{mathrsfs}
\usepackage{algorithm,algorithmic}
\usepackage{boxedminipage}
\usepackage{pgf,pgfarrows}
\usepackage{subfigure}
\usepackage{xcolor}
\usepackage{siunitx}
\usepackage{hyperref}
\usepackage[hyperpageref]{backref}
\usepackage{booktabs}
\usepackage{colortbl}

\graphicspath{{fig/}}

\hypersetup{
pdfauthor={Andreas Mang and Lars Ruthotto},
linkcolor=red,
citecolor=green}

\newcommand{\R}{\mathbb{R}}

 \renewcommand{\grad}{\nabla}

\renewcommand{\phi}{\mathbf{\varphi}}

\newcommand{\bF}{\mathbf{F}}

\newcommand{\CC}{\mathcal{C}}

\newcommand{\CT}{\mathcal{T}}
\newcommand{\CR}{\mathcal{R}}
\newcommand{\CO}{\mathcal{O}}

\newcommand{\CJ}{\mathcal{J}}

\newcommand{\CD}{\mathcal{D}}

\newcommand{\CS}{\mathcal{S}}                          

\newcommand{\bfF}{\mathbf{F}}

\newcommand{\bfA}{\mathbf{A}}

\newcommand{\bfI}{\mathbf{I}}
\newcommand{\bfD}{\mathbf{D}}

\newcommand{\bfH}{\mathbf{H}}
\newcommand{\bfB}{\mathbf{B}}
\newcommand{\bfx}{\mathbf{x}}

\newcommand{\bfu}{\mathbf{u}}
\newcommand{\bfy}{\mathbf{y}}

\newcommand{\bfL}{\mathbf{L}}

\newcommand{\bfv}{\mathbf{v}}

\newcommand{\x}{x}														
\renewcommand{\xi}[1]{\x_{#1}}								

\newcommand{\hf}{\frac{1}{2}}

\newcolumntype{R}{>{\columncolor{gray!20}}r}
\newcolumntype{L}{>{\columncolor{gray!20}}l}

\newcounter{runidnum}
\newcommand{\runid}{\stepcounter{runidnum}\#\therunidnum}

\newdimen\iwidth
\newdimen\iheight

\title{A Lagrangian Gauss--Newton--Krylov Solver for Mass- and Intensity-Preserving Diffeomorphic Image Registration}
\author{Andreas Mang\thanks{Institute for Computational Engineering and Sciences, University of Texas at Austin, Austin, Texas, USA. ({\tt andreas@ices.utexas.edu})} \and Lars Ruthotto\thanks{Department of Mathematics and Computer Science, Emory University, Atlanta, Georgia, USA. ({\tt lruthotto@emory.edu})}}

\begin{document}

\maketitle

\begin{abstract}
We present an efficient solver for diffeomorphic image registration problems in the framework of \emph{Large Deformations Diffeomorphic Metric Mappings} (LDDMM). We use an optimal control formulation, in which the velocity field of a hyperbolic PDE needs to be found such that the distance between the final state of the system (the transformed/transported template image) and the observation (the reference image) is minimized. Our solver supports both stationary and non-stationary (i.e., transient or time-dependent) velocity fields. As transformation models, we consider both the transport equation (assuming intensities are preserved during the deformation) and the continuity equation (assuming mass-preservation).

We consider the reduced form of the optimal control problem and solve the resulting unconstrained optimization problem using a discretize-then-optimize approach. A key contribution is the elimination of the PDE constraint using a Lagrangian hyperbolic PDE solver. Lagrangian methods rely on the concept of characteristic curves. We approximate these curves using a fourth-order Runge-Kutta method. We also present an efficient algorithm for computing the derivatives of the final state of the system with respect to the velocity field. This allows us to use fast Gauss-Newton based methods. We present quickly converging iterative linear solvers using spectral preconditioners that render the overall optimization efficient and scalable. Our method is embedded into the image registration framework FAIR and, thus, supports the most commonly used similarity measures and regularization functionals. We demonstrate the potential of our new approach using several synthetic and real world test problems with up to 14.7 million degrees of freedom.
\end{abstract}



\section{Introduction} 
\label{sec:introduction}

In this paper, we present efficient numerical methods for diffeomorphic image registration in the framework of \emph{Large Deformation Diffeomorphic Metric Mapping}~({\bf LDDMM})~\cite{Trouve1998,DupuisEtAl1998,BegEtAl2005}. We use an optimal control formulation similar to the one in~\cite{BorziEtAl2002a,BorziEtAl2002b,BenziEtAl2011,MangBiros2015IS,MangBiros2016IS,MangBiros2016SL}. Here, the task is to find a smooth velocity field $v$ such that the distance between two images (or densities), $\CT$ (the \emph{template image}) and $\CR$ (the \emph{reference image}), is minimized, subject to some regularization norm for $v$ and a transformation model, given by a hyperbolic partial differential equation ({\bf PDE}) that models the deformation of $\CT$. We consider the \emph{transport equation} (assuming intensities are related at corresponding points) and the \emph{continuity equation} (assuming that mass is preserved) as constraints. The connection to traditional image registration formulations~\cite{Modersitzki2004,Modersitzki2009} is that a sufficiently smooth velocity field $v$ gives rise to a diffeomorphism $y$ via the method of characteristics.
Vice versa, representing diffeomorphisms through velocity fields has been used for efficient statistical analysis; see, e.g.,~\cite{ArsignyEtAl2006}.

Solving the variational problem associated with LDDMM is, in theory, known to yield a diffeomorphic transformation $y$ if $v$ is sufficiently smooth~\cite{DupuisEtAl1998,Trouve1998,BegEtAl2005}. Although, the theory of diffeomorphic registration using LDDMM is well explored~\cite{Miller:2001a,Younes:2007a,Younes:2009a}, efficient numerical optimization is not. Until recently~\cite{AshburnerFriston2011,Hernandez2014,MangBiros2015IS,MangBiros2016IS,MangBiros2016SL} mostly first-order optimization methods were used; see, e.g.,~\cite{BegEtAl2005,Vialard:2011ef,ChenLorenz2011,PolzinEtAl2016}. A key component in LDDMM is the numerical method for solving the hyperbolic PDE. Hyperbolic PDE solvers can be roughly divided into Eulerian (in which the density is discretized at the same locations for each time point) and Lagrangian (in which the grid moves over time along the characteristic curves) solvers; see also~\cite{LeVeque1992,EwingWang2001,LeVeque2002}. Intermediates are Semi-Lagrangian~(\textbf{SL}) methods, which follow the velocity for a short time step and then estimate the density at fixed points. In this work, we use a Lagrangian solver.

It is well known that explicit Eulerian methods for hyperbolic PDEs require the size of the time step to be sufficiently small to ensure numerical stability. The maximal admissible time step size depends on the accuracy of the spatial discretization and the magnitude of the velocities. In optimal control problems, like ours, the velocity field $v$ is not known a priori and, thus, it is difficult to come up with an efficient and stable choice of the time step. SL and Lagrangian solvers are explicit methods that, unlike explicit Eulerian methods, are stable without a restriction on the maximal admissible time step size. One drawback of SL methods is their memory requirements. For efficient derivative (or sensitivity) computations in Gauss--Newton type optimization schemes, we have to store intermediate images~\cite{MangEtAl2016SC}. Further, SL methods require a repeated interpolation of the initial image and may therefore introduce severe dissipation if implemented naively.\footnote{We note, that interpolation errors and numerical diffusion can be minimized, by, e.g., applying high-order interpolation schemes and/or evaluating the interpolation on a finer grid; this is costly.} As we will see, Lagrangian methods require only one image interpolation at the final time. Secondly, derivatives of the Lagrangian solver can be obtained efficiently, without storing intermediate variables. A feature of SL and Lagrangian methods is that they can be easily modified to solve intensity- and mass-preserving problems, since the characteristic curves coincide.

The key idea of our work is to use a \emph{discretize-then-optimize} strategy based on a Lagrangian hyperbolic PDE solver to efficiently solve the reduced formulation of the PDE-constrained optimization problem arising in LDDMM. We show that Lagrangian methods lead to a finite dimensional optimization problem that can be solved using an inexact Gauss--Newton method. Our PDE solver requires numerical computation of the characteristic curves. We use a fourth-order Runge--Kutta ({\bf RK4}) method to numerically approximate the transformation $y$ that is associated with a, in general non-stationary, velocity field $v$. As we show, derivatives of the transformation with respect to $v$ can be derived analytically and computed efficiently. Due to the hyperbolic nature of the PDE, the derivatives can be represented as sparse matrices; the procedure can be paralellelized: characteristics starting at different points can be computed independently. Given these characteristics, the hyperbolic PDEs can be solved by a single interpolation step (for the advection equation) or the particle-in-cell method (for the continuity equation).

\subsection{Contributions} 
\begin{itemize}
\item We propose a discretize-optimize method for solving LDDMM using a Lagrangian hyperbolic PDE solver. Our scheme is based on an RK4 method to approximate the characteristic curves and we derive an efficient algorithm for computing the derivative of the solver with respect to the velocities.
\item The storage requirement of our method is independent on the number of time steps used in the numerical solver. Also, the Hessian of the objective function can be build explicitly at moderate costs, which is useful, e.g., to accelerate matrix-vector products. In this work, we use and numerically study the convergence of spectral preconditioners to iteratively solve the Gauss--Newton system.
\item We extend the LDDMM framework to mass-preserving registration, which has been proved to be an adequate model for many relevant biomedical applications involving with density images, e.g., in~\cite{ChangFitzpatrick1992,RehmanEtAl2009,DawoodEtAl2010,GigengackEtAl2012,RuthottoEtAl2012EPI}.
\item We derive a flexible framework supporting both stationary and non-stationary velocity fields. Our methods are embedded into the FAIR framework~\cite{Modersitzki2009}. This allows us to consider different regularization norms and distance measures. Our implementation is freely available as an add-on to FAIR at:

\begin{center}
	\url{https://github.com/C4IR/FAIR.m/tree/master/add-ons/LagLDDMM}
\end{center}

\item We provide detailed numerical experiments on four different data sets that demonstrate the flexibility and effectiveness of our method. We show that our prototype implementation is competitive to state-of-the-art packages for diffeomorphic image registration~\cite{BurgerEtAl2013, RuthottoGreifModersitzki2016}. We study registration quality for synthetic benchmark problems and real-world applications leading to optimization problem with up to 14.7 million degrees of freedom.

\end{itemize}

\subsection{Related Work} 

We limit this review to work closely related to ours. For a general insight into the area of image registration, its applications, and its formulation we refer to~\cite{Modersitzki2004,Modersitzki2009,Sotiras:2013a}. Our work builds upon the LDDMM framework described in~\cite{DupuisEtAl1998,Trouve1998,BegEtAl2005}, which is based on the pioneering work on velocity-based fluid registration described in~\cite{Christensen:1996a}. We adopt an optimal control point of view; we also do not directly invert for the transformation $y$ but for its velocity $v$. We arrive at a hyperbolic PDE-constrained optimization problem. We refer to~\cite{Gunzburger:2003a,Hinze:2009a,Herzog:2010a,Borzi:2012a} for a general introduction into optimal control theory and developments in PDE-constrained optimization. Related optimal control formulations for diffeomorphic image registration can, e.g., be found in~\cite{BorziEtAl2002a,BorziEtAl2002b,HartEtAl2009,BenziEtAl2011,ChenLorenz2011,MangBiros2015IS,MangBiros2016IS,MangBiros2016SL,MangEtAl2016SC}. Other formulations for velocity-based diffeomorphic image registration are described in~\cite{Ashburner2007,Vercauteren:2009a,AshburnerFriston2011}. Our work also shares characteristics with optical flow formulations~\cite{BorziEtAl2002a,BorziEtAl2002b,ChenLorenz2011}. Our formulation for mass-preserving registration problems is related to the Monge--Kantorovich functional arising in optimal mass transport~\cite{BenziEtAl2011,HakerEtAl2004}.

Most work on large deformation diffeomorphic image registration still considers first-order methods for numerical optimization (see, e.g.,~\cite{AvantsEtAl2006,AvantsEtAl2008,AvantsEtAl2011,BorziEtAl2002a,BegEtAl2005,HartEtAl2009,Lee:2010a,ChenLorenz2011,Vialard:2011ef}); the exceptions are~\cite{AshburnerFriston2011,MangBiros2015IS,MangBiros2016IS,MangBiros2016SL,Vercauteren:2009a}. First-order schemes for numerical optimization do in general require a larger number of iterations than Newton type optimization schemes.\footnote{We note that in LDDMM most implementations use a gradient descent scheme in the Sobolev space induced by the regularization operator (dual space). This leads to a significant speedup compared to standard gradient descent approaches. However, it has been demonstrated experimentally that Gauss--Newton--Krylov methods are superior~\cite{MangBiros2015IS}.} The work in~\cite{AshburnerFriston2011} uses geodesic shooting and estimates the initial value of a non-stationary velocity field that parameterizes the diffeomorphism $y$. Other approaches that reduce the size of the optimization problem are based on stationary velocity fields; see, e.g.,~\cite{HernandezEtAl2009,LorenziEtAl2013,MangBiros2016IS,MangBiros2016SL,MangEtAl2016SC}.

PDE-constrained optimization commonly requires a repeated solution of the forward problem. Thus, the design of an efficient forward solver is critical. The approaches described in~\cite{BenziEtAl2011,ChenLorenz2011,MangBiros2015IS,MangBiros2016IS,MangBiros2016SL,HartEtAl2009} are based on an Eulerian formulation. They employ explicit high-order schemes~\cite{BorziEtAl2002a,BorziEtAl2002b,MangBiros2015IS,MangBiros2016IS,HartEtAl2009,PolzinEtAl2016}, which suffer from a restriction on a maximally admissible time step, implicit schemes~\cite{BenziEtAl2011}, or explicit SL schemes~\cite{MangBiros2016SL,MangEtAl2016SC,ChenLorenz2011}. The latter were originally proposed in the context of weather prediction~\cite{StaniforthCote1991}. They are a hybrid between Lagrangian and Eulerian schemes, and unconditionally stable. SL schemes have been used in the context of Lagrangian formulations for diffeomorphic image registration (to compute the characteristics)~\cite{BegEtAl2005,HernandezEtAl2009}. Conditionally stable schemes require small time steps, which can result in a significant amount of memory that needs to be allocated to store the time-space fields necessary to evaluate the gradient or Hessian. This makes a direct application of these type of methods to large-scale 3D problems challenging. One remedy is to turn to parallel architectures and use sophisticated checkpointing schemes in time to reduce the amount of memory that has to be allocated; see, e.g.,~\cite{Akcelik:2002a}. Implicit schemes typically suffer from severe numerical diffusion. The same is true for straightforward implementations of SL schemes. Pure Lagrangian schemes for diffeomorphic image registration have been described in~\cite{AvantsEtAl2006,AvantsEtAl2008,AvantsEtAl2011}. The time integration for computing the characteristics in~\cite{AvantsEtAl2006,AvantsEtAl2008,AvantsEtAl2011} is based on a first-order explicit scheme.

What sets our work apart is the \emph{numerical solver} and the generalization of our formulation to both the \emph{transport} and the \emph{continuity equation}. Most existing works on optimal control formulations for diffeomorphic image registration consider an optimize-then-discretize approach~\cite{ChenLorenz2011,MangBiros2015IS,MangBiros2016IS,MangBiros2016SL,MangEtAl2016SC}. We use a discretize-then-optimize strategy instead.\footnote{Advantages and disadvantages of these two techniques are discussed, e.g., in~\cite{Gunzburger:2003a}.} Similar to~\cite{MangBiros2015IS}, we describe a method that can handle stationary and non-stationary velocity fields. Our numerical scheme is, likewise to~\cite{AvantsEtAl2006,AvantsEtAl2008,AvantsEtAl2011}, based on a purely Lagrangian approach. We consider a reduced formulation of the PDE-constrained optimization problem arising in LDDMM, i.e., we eliminate the hyperbolic PDE constraint (state equation) from the variational problem. We use a Lagrangian solver to parameterize the final state in terms of the velocity. In doing so we avoid many of the complications we reviewed above: our method is unconditionally stable, limits numerical diffusion, and does not require the storage of multiple space-time fields for the evaluation of the gradient or Hessian operators. The work in~\cite{AvantsEtAl2006,AvantsEtAl2008,AvantsEtAl2011} uses first-order information for numerical optimization and a first-order accurate explicit time integrator to compute the characteristic. We use a fourth-order, explicit RK scheme instead. We derive expressions for the exact derivative of the characteristics. We arrive at a Gauss--Newton--Krylov scheme that---combined with an efficient iterative linear solver---yields an approximate solution within a few iterations, and has an overall algorithmic complexity of $\CO(n \log(n))$, where $n$ is the dimension of the discretized velocity field (i.e., the number of unknowns).


\begin{table}
\caption{Commonly used symbols and abbreviations.}
\scriptsize\centering
\begin{tabular}{ll}\toprule
DCT & discrete cosine transform\\
FAIR & Flexible Algorithms for Image Registration~\cite{Modersitzki2009}\\
LDDMM & large deformation diffeomorphic metric mapping\\
MRI & magnetic resonance imaging\\
PDE & partial differential equation\\
PET & positron emission tomography\\
PIC & particle-in-cell (method)\\
SSD & sum-of-squared-differences
\\\midrule
$\CR(x)$ & reference / fixed image \\
$\CT(x)$ & template image (image to be registered) \\
$\CC$ & PDE constraint \\
$\CD$ & distance or similarity measure\\
$\CS$ & regularization model (smoother)\\
$\alpha$ & regularization weight\\
$x$ & spatial coordinate; $x\in\Omega$\\
$\Omega$ & spatial domain; $\Omega\subset\R^d$\\
$v(x,t)$ & velocity field\\
$u(x,t)$ & transported image intensities\\
$y(x)$ & transformation / mapping\\
$N$ & number of time steps for computing the characteristic\\
$n$ & number of unknowns (i.e., the dimension of the discretized velocity field)\\
$n_t$ & number of cells in space-time grid\\
$I$ & interpolation operator\\
$\nabla$ & gradient operator\\
$\nabla\cdot$ & divergence operator\\
$\partial_t$ & time derivative
\\\bottomrule
\end{tabular}
\end{table}


\section{Mathematical Formulation} 
\label{sec:mathematical_formulation}

We describe the variational optimal control formulation of the LDDMM problem next. We denote the image domain by $\Omega \subset \R^d$, where $d\in\{2,3\}$ represents the spatial dimension. We assume that the \emph{template} and the \emph{reference image}, denoted by $\CT : \Omega \to \R$ and $\CR: \Omega \to \R$, are compactly supported on $\Omega$ and continuously differentiable. Given these two images, the task of image registration is to find a \emph{plausible} transformation $y: \Omega \to \R^d$ so that the transformed template image $\CT\circ y$ becomes \emph{similar} to the reference image $\CR$~\cite{Modersitzki2009}. The definitions of \emph{plausibility}, \emph{similarity}, and the \emph{transformation model} depend on the context; see~\cite{Modersitzki2009,Modersitzki2004,Sotiras:2013a} for examples. Many relevant applications, e.g., in medical imaging, require that plausible transformations are diffeomorphic, i.e., smooth mappings with a smooth inverse. One framework that contains the most commonly used definitions of these three terms within the medical imaging application domain is LDDMM~\cite{DupuisEtAl1998,Trouve1998,BegEtAl2005}. The variational optimal control formulation of the LDDMM problem can, in general format, be stated as follows:
\begin{equation}
\label{eq:variational}
\min_{v,u}\left\{ \CJ(v,u) :=  \CD(u(\cdot,1), \CR) + \alpha \CS(v) \right\} \quad \text{ subject to }\quad \quad \CC(v,u) = 0,
\end{equation}
\noindent where $\CD$ is a distance (or similarity) measure, $\CS$ is a regularizer (smoother), $v : \Omega \times [0,1] \to \R^d$ is the sought after velocity field, and $u : \Omega \times [0,1] \to \R$ is a time series of images. In an optimal control context, $v$ is commonly referred to as the \emph{control variable} and $u$ as the \emph{state variable}. Here, $\alpha>0$ is a regularization parameter that balances between minimizing the image distance and the smoothness of the velocity field (and consequently controls the properties of the resulting transformation). In the current work, we assume that $\alpha$ is chosen by the user and refer to~\cite{Hansen1998,HaberOldenburg2000,Vogel2002} for some works on automatic selection of the parameters.\footnote{Examples for an automatic selection of the regularization parameter in the context of image registration can, e.g., be found in~\cite{Haber:2006a,MangBiros2015IS}.}

In this work, we assume that the constraint $\CC$, which describes the transformation model, is given either by the \emph{advection} (also called \emph{transport}) equation
\begin{equation}
\label{eq:transport}
\CC(u,v) = \left\{
\begin{array}{l}
\partial_t u(x,t) + v(x,t) \cdot \grad u(x,t) = 0,\\
u(x,0) = \CT(x)
\end{array}\right.
\end{equation}

\noindent or the \emph{continuity} equation
\begin{equation}
\label{eq:continuity}
\CC(u,v) = \left\{ \begin{array}{l}
\partial_t u(x,t) + \grad \cdot(u(x,t) v(x,t)) = 0,\\
u(x,0) = \CT(x).
\end{array}\right.
\end{equation}

\noindent The former assumes intensity values are preserved during the transformation; the latter preserves the overall mass of the image. The choice of the transformation model depends on the application. Intensity preservation is commonly used, e.g., for registration of medical images acquired from different subjects~\cite{OuEtAl2014}. Mass-preservation has been successfully used, e.g., for motion correction in position emission tomography~({\bf PET})~\cite{DawoodEtAl2010,GigengackEtAl2012} or artifact correction of magnetic resonance imaging ({\bf MRI})~\cite{ChangFitzpatrick1992, RuthottoEtAl2012EPI}.

The models in~\eqref{eq:transport} and~\eqref{eq:continuity} can be used to establish point-to-point correspondences between the template and the reference image. One way of showing this is the method of characteristics~\cite{LeVeque2002,Evans2010}. To better illustrate this, we consider the advection equation~\eqref{eq:transport}, for which the intensity $u$ is constant along the characteristics. This means that, for all $y_0 \in \Omega$ and $t \in [0,1]$, it holds that $u(y(v,y_0,0,t),t) = \CT(y_0)$ where the characteristic curve $t \mapsto y(v,y_0,0,t)$ satisfies
\begin{equation}
\label{eq:characteristic}
\partial_t y(v,y_0,0,t) = v( y(v,y_0,0,t),t)
\quad \text{ and } \quad y(v,y_0,0,0) = y_0.
\end{equation}

\noindent Similarly, the characteristics can be traced backwards in time to compute the state at some point $y_1\in\R^d$ at $t=1$. The position of the point is given by $t \mapsto y(v,y_1,1,t)$, which satisfies~\eqref{eq:characteristic} with final time condition $y(v,y_1,1,1)=y_1$. Clearly, both operations are inverse to one another. That is, for all $x\in\R^n$ it holds that $y(v,y(v,x,0,1),1,0) = x$, i.e., a composition of both maps yields identity.

Note that~\eqref{eq:transport} through~\eqref{eq:characteristic} involve a non-stationary (i.e., time-dependent) velocity field $v$. This is a key assumption in the original LDDMM formulation~\cite{BegEtAl2005}. To make the problem computational tractable it is, however, often assumed that $v$ is stationary (\emph{stationary velocity field} based registration)~\cite{ArsignyEtAl2006,Ashburner2007,HernandezEtAl2009,MangBiros2016IS,MangEtAl2016SC,Pai:2016a}.\footnote{Another strategy to reduce the computational burden is to invert for an initial momentum~\cite{Vialard:2011ef} that encodes the trajectory of the diffeomorphism.} The numerical framework we propose in this paper can efficiently handle both stationary and non-stationary velocities. As demonstrated in our numerical experiments in Sec.~\ref{sub:2d_lddmm_example},  the stationary model is less flexible in that we can only invert for a subset of the deformation maps living on the manifold of diffeomorphisms. Our experiments  also suggest that stationary velocity fields are adequate for registration problems involving two topologically similar images, yielding little to no difference in the recovered deformation map~\cite{Ashburner2007,HernandezEtAl2009,MangBiros2015IS}. However, using non-stationary velocity fields may become critical in applications involving large and highly nonlinear transformations and/or the registration of time series of images with large motion between time frames (e.g., typically seen in tracking or optical flow problems).

\subsection{Regularization functionals} 
\label{sub:regularization_functionals}

Due to the ill-posedness of the image registration problem, the literature on regularization in image registration is rich; see, e.g.,~\cite{Modersitzki2004,FischerModersitzki2008,Modersitzki2009,Sotiras:2013a} for extensive overviews. It is established that the existence and regularity of a diffeomorphic map $y$ depends on the smoothness of the velocity field $v$ as well as the smoothness of the images $\CR$ and $\CT$~\cite{DupuisEtAl1998,Trouve1998,MangBiros2016IS,ChenLorenz2011}. Modeling the images as functions of bounded variation, $H^3$-regularity~\cite{ChenLorenz2011} is required (assuming that $v$ is divergence free). For continuous images we can relax the $H^3$-regularity to an $H^2$-regularity~\cite{BegEtAl2005} or---under additional assumptions on the divergence of $v$---even to an $H^1$-regularity~\cite{ChenLorenz2011}.\footnote{We use interpolation and padding of the discrete image data to obtain continuously differentiable and compactly supported functions (see Sec.~\ref{sec:discretization_with_lagrangian_methods}).} Most implementations for LDDMM consider an $H^2$-norm for $\CS$ in~\eqref{eq:variational} (or an approximation based on a Gaussian kernel within a gradient descent scheme in the Sobolev space induced by the regularization norm); see, e.g.,~\cite{AshburnerFriston2011,BegEtAl2005,HernandezEtAl2009}.

In our framework, we regard the regularizer as a modular component that can be replaced or extended. In practical applications we control the regularization parameter by monitoring the Jacobian in an attempt to generate transformations that are  diffeomorphic (in a discrete setting) and yield high-fidelity (low mismatch) results. In our numerical examples, we consider $H^1$- and $H^2$-seminorms as regularization models. These type of regularization models are also referred to as \emph{diffusive} or \emph{curvature} regularizers in the context of traditional variational image registration formulations~\cite{FischerModersitzki2003,Modersitzki2004,Modersitzki2009}, respectively. Assuming that $v$ is non-stationary, we have
\begin{equation*}
\CS^{\mathrm{diff}}(v) = \hf \int_0^1 \int_{\Omega} \sum_{k=1}^d | \nabla v^k(x,t)|^2 \mathrm{d}x \mathrm{d}t
\end{equation*}

\noindent and
\begin{equation*}
\CS^{\text{curv}}(v)  = \hf \int_0^1 \int_{\Omega} \sum_{k=1}^d | \Delta v^k(x,t)|^2 \mathrm{d}x \mathrm{d}t,
\end{equation*}

\noindent respectively. Regularization of stationary velocity fields is along the same lines, by simply dropping the time integration in the above equations. Our formulation can also be extended to enforce smoothness in time, as also done in~\cite{BorziEtAl2002a}.

\section{Numerical Methods} 
\label{sec:discretization_with_lagrangian_methods}

In this section, we describe a discretize-then-optimize approach for solving the variational problem~\eqref{eq:variational}. We eliminate the hyperbolic PDE constraints~\eqref{eq:transport} and~\eqref{eq:continuity} using Lagrangian methods. We then describe the discretization of the objective functional itself. Following~\cite{Modersitzki2009}, we consider a multilevel Gauss--Newton method that allows us to efficiently solve the discrete optimization problem. Finally, we give some details about our implementation as an extension to the FAIR toolbox~\cite{Modersitzki2009}.

Assume, for simplicity, that the domain $\Omega = (0,1)^d$ is divided into a regular mesh of $m$ cells of edge length $h=1/m$ along each coordinate direction. We use interpolation and padding of the discrete image data to obtain continuously differentiable and compactly supported functions. We approximate integrals in~\eqref{eq:variational} by a midpoint quadrature rule, which requires evaluating the final state on the cell-centered points, $\bfx_c \in \R^{d\cdot m^d}$, of the grid. Without loss of generality, we assume that the velocity field $v$ is discretized on the same mesh. However, the domain size and number of cells can be varied in practice. The discrete velocity field, denoted by $\bfv \in \R^n$, is discretized in time at the nodes of a regular grid with $n_t$ cells and in space at cell-centered grid points. The total number of unknowns is $n = (n_t+1) \cdot d \cdot(m^d)$.

\subsection{Lagrangian Methods for Hyperbolic PDEs} 
\label{sub:lagrangian_methods_for_hyperbolic_pdes}

Lagrangian methods exploit the fact that solutions to hyperbolic PDEs evolve along characteristic curves~\cite{LeVeque2002,Evans2010}. These methods are Lagrangian in the sense that the transport of the density is referred to in the moving coordinate system. Lagrangian methods typically consist of two steps: First, the characteristics are computed numerically. Then, in a second step, the final image (or density) is computed. While the characteristic curves are identical for the advection and the continuity equation, the computation of the second part is not.

\paragraph{Step 1 (Computing the Characteristics)} 
\label{par:computing_the_characteristics}

Here, we describe our implementation for computing the characteristic curve passing through a given point. The velocity field $v : \Omega \times [0,1] \to \R^d$ is known and represented by the coefficients $\bfv \in \R^n$. We solve~\eqref{eq:characteristic} using an RK4 method with $N$ equidistant time steps of size $\Delta t = \pm 1/N$; the sign of the time step depends on whether the characteristics are computed forward or backward in time. Note that the number of time steps $N$ for the RK4 method and the number of cells $n_t$ in the space-time grid do not necessarily have to be equal. The former is a parameter of the numerical solver and controls the accuracy of the characteristics. The latter is a modeling parameter and ultimately controls the search space for the transformation $y$.

Our choice of an RK4 scheme is motivated by accuracy considerations for computing the characteristics. For simplicity, we illustrate the concept of integrating $v$ based on a first-order forward Euler scheme. The derivation of our RK4 scheme is along the same lines; it is outlined in Algorithm~\ref{alg:rk4}.

Let $\bfx \in \R^{d\cdot n_p}$ be the coordinates of the start (or end) points of the characteristics, e.g., the cell-centers of a regular mesh. Introducing the (time-dependent) transformation $y: \R^n \times \R^{d\cdot n_p} \times [0,1]^2 \to \R^{d\cdot n_p}$, and imposing the initial condition $y(\bfv,\bfx,0,0) = \bfx$, we compute
\begin{equation}
\label{eq:forwardEuler}
y(\bfv,\bfx,0,t_{k+1}) =  y(\bfv,\bfx,0,t_k) + \Delta t\ I( \bfv,y(\bfv,\bfx,0,t_k), t_k ),  \quad \forall\ k=0,1,\ldots,N-1,
\end{equation}

\noindent where $t_k = k\Delta t$ are the time points and $I$ interpolates the velocity field $\bfv$ at the transformed points $y$. In our experiments, we use a bi- or trilinear interpolation model in space and a linear interpolation model in time, applied separately to each component of the velocity field $\bfv$. We found by experimentation that using low-order interpolation schemes for the (smoothness regularized) velocity fields is sufficiently accurate for our numerical scheme to yield high-fidelity results in practical applications. We note that the interpolation is a modular component; it can be replaced by (computationally more expensive) higher-order methods. Notice, that the positions at previous time steps, $y(\bfv,\bfx,0,0),\ldots,y(\bfv,\bfx,0,t_{k-1})$, do not enter the computation in~\eqref{eq:forwardEuler}; the memory requirements of our numerical scheme are independent of the number of time steps $N$. The end points of the characteristics are then $y(\bfv,\bfx,0,1)$, which, if $\bfx = \bfx_c$, can also be interpreted and visualized as a deformed regular grid.

\begin{algorithm}
\caption{RK4 method for computing the characteristics $\bfy$ and the derivative $d_{\bfv}\bfy$.}
\label{alg:rk4}
\begin{algorithmic}
\STATE \textbf{Input:} Discrete (non-stationary) velocity field $\bfv \in \R^n$, start points of characteristics $\bfx \in \R^{d\cdot n_p}$, number of time steps $N$
\STATE \textbf{Set:} $\bfy \leftarrow \bfx$, $ d_{\bfv} \bfy \leftarrow 0$, $\Delta t \leftarrow 1/N$.
\FOR{$k=0,1,\ldots,N-1$}
\STATE compute $\bfv_1 \leftarrow I\big(\bfv,\bfy,0,t_k\big)$ (spatio-temporal vector field interpolation) and set $\bfy_1 \leftarrow \bfy +  (\Delta t/2) \bfv_1$
\STATE compute $\bfv_2 \leftarrow I\big(\bfv,\bfy_1,0,t_{k+1/2}\big)$  and set $\bfy_2 \leftarrow \bfy + (\Delta t/2) \bfv_2$
\STATE compute $\bfv_3 \leftarrow I\big(\bfv,\bfy_2,0,t_{k+1/2}\big)$  and set $\bfy_3 \leftarrow \bfy + (\Delta t/2) \bfv_3$
\STATE compute $\bfv_4 \leftarrow I\big(\bfv,\bfy_3,0,t_{k+1}\big)$
\IF{derivative required}
	\STATE  $ \bfD_1 \leftarrow  d_{\bfv} I\big(\bfv,\bfy,0,t_k\big) + d_\bfy I\big(\bfv,\bfy,0,t_k\big) \ d_{\bfv} \bfy$
	\STATE  $ \bfD_2 \leftarrow d_{\bfv} I\big(\bfv,\bfy_1,0,t_{k+1/2}\big) + d_\bfy I\big(\bfv,\bfy_1,0,t_{k+1/2}\big) \big(d_{\bfv} \bfy + (\Delta t/2) \bfD_1\big)$
	\STATE  $ \bfD_3 \leftarrow d_{\bfv} I\big(\bfv,\bfy_2,0,t_{k+1/2}\big) + d_\bfy I\big(\bfv,\bfy_2,0,t_{k+1/2}\big) \big(d_{\bfv} \bfy + (\Delta t/2)\bfD_2\big)$
	\STATE  $ \bfD_4 \leftarrow  d_{\bfv} I\big(\bfv,\bfy_3,0,t_{k+1}\big)+ d_\bfy I(\bfv,\bfy_3,0,t_{k+1}) \big(d_{\bfv} \bfy + \Delta t \bfD_3\big)$
	\STATE $d_{\bfv}\bfy \leftarrow d_{\bfv}\bfy + (\Delta t/6) \big(\bfD_1 + 2 \bfD_2 + 2 \bfD_3 + \bfD_4 \big)$
\ENDIF
\STATE $\bfy \leftarrow \bfy + (\Delta t / 6) \big(\bfv_1 + 2 \bfv_2 + 2 \bfv_3 + \bfv_4 \big)$
\ENDFOR
\STATE \textbf{Output:} end of characteristics, $\bfy\in\R^{d\cdot n_p},$ and (if required) gradient, $d_{\bfv} \bfy \in\R^{n \times d \cdot n}$
\end{algorithmic}
\end{algorithm}


\paragraph{Step 2 (Solving the hyperbolic PDE)} 
\label{par:eliminating_the_constraint}

\begin{figure}
\centering
\begin{tabular}{cc}
\begin{tikzpicture}[scale=.6]
   \filldraw [red!40] (-1,-1) circle (6pt);
   \filldraw [red!40] (-1,1) circle (6pt);
   \filldraw [red!40] (1,-1) circle (6pt);
   \filldraw [red!40] (1,1) circle (6pt);

   \draw [->,gray, ultra thick](-1,-1) to (-0.45,-0.85);
   \draw [->,gray, ultra thick](-1,1) to (-0.45,-0.55);
   \draw [->,gray, ultra thick](1,-1) to (-0.2,-0.8);
   \draw [->,gray, ultra thick](1,1) to (-0.2,-.55);
   \draw [->,blue, ultra thick](-0.18,-0.58)  to[in=180,out=20]  (2.8,1);
 \filldraw [red] (-0.3,-0.7) circle (6pt);
 \filldraw [red] (3,1) circle (6pt);

 \pgftext[base,x=3.5cm,y=0.7cm] {\LARGE $x$};
 \pgftext[base,x=-0.2cm,y=-1.8cm] {\LARGE $y(\bfv,x,1,0)$};

  \draw[step=2cm,black!40] (-3,-2) grid (7,5.8);

   \end{tikzpicture} & \begin{tikzpicture}[scale=.6]

   \filldraw [red!40] (3,3) circle (6pt);
   \filldraw [red!40] (3,5) circle (6pt);
   \filldraw [red!40] (5,3) circle (6pt);
   \filldraw [red!40] (5,5) circle (6pt);

  \draw [->,gray, ultra thick] (4.2,3.4) to (3,3) ;
   \draw [->,gray, ultra thick](4.2,3.4) to (3,5)  ;
   \draw [->,gray, ultra thick](4.2,3.4)   to (5,3)  ;
   \draw [->,gray, ultra thick](4.2,3.4)   to (5,5)   ;
   \draw [->,blue, ultra thick] (3,1) to[out=10,in=-80] (4.2,3.4);
   \filldraw [red] (4.2,3.6) circle (6pt);
   \filldraw [red] (3,1) circle (6pt);

 \pgftext[base,x=4.0cm,y=0.8cm] {\LARGE $x$};
 \pgftext[base,x=6.2cm,y=3.5cm] {\LARGE $y(\bfv,x,0,1)$};

  \draw[step=2cm,black!40] (-3,-2) grid (7,5.8);

   \end{tikzpicture}       \\
advection; $\CT(y(\bfv,x,1,0)))$      & continuity; $\bfF(y(\bfv,x,0,1)) \CT(x)$
\end{tabular}
\caption{Illustration of Lagrangian methods for solving linear hyperbolic PDEs. In both cases, the characteristic curves (indicated by a blue line) are computed starting from a grid point $x$. Left: The advection problem is solved by traveling along the characteristics backwards in time to the non-grid point $y(\bfv,x,1,0)$. The associated image intensity is computed by interpolating the intensities of the adjacent cells. Right: The continuity equation is solved by pushing the mass $\CT(x)$ from $x$ along the characteristics to the non-grid point $y(\bfv,x,0,1)$ and then distributing $\CT(x)$ among the cells adjacent to $y(\bfv,x,0,1)$.}
\label{fig:lagrangian}
\end{figure}
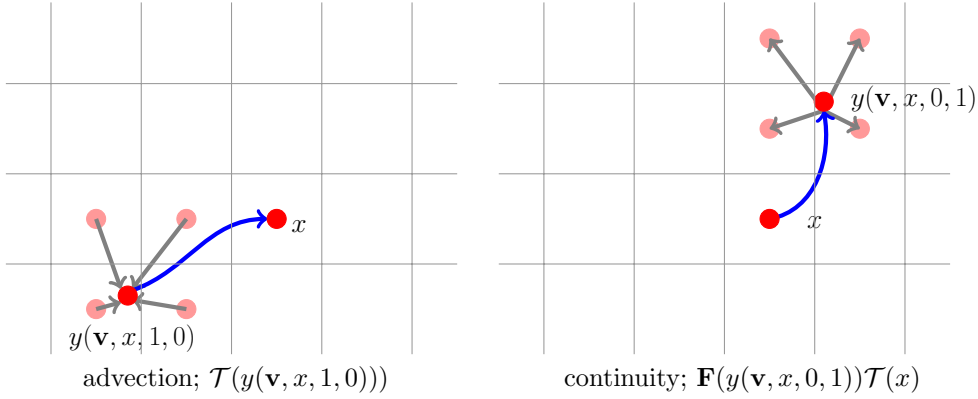

While solutions to both the transport and the continuity equations evolve along the same characteristics, the steps for computing the transported quantity at $t=1$ vary. Thus, we discuss both cases separately. An illustration of both schemes can be found in Fig.~\ref{fig:lagrangian}.

\emph{Transport equation}: Considering the transport equation~\eqref{eq:transport}, we compute the intensities of the advected image $\CT$ on the deformed cell-centered grid $\bfx_c$ by following the characteristics backwards in time. This yields
\begin{equation}
\label{eq:rhoEndIR}
u(\bfx_c,1) = \CT(y(\bfv,\bfx_c,1,0)).
\end{equation}

In general, $y(\bfv,\bfx_c,1,0)$ does not coincide with a grid point; the intensity has to be computed by interpolation. In our numerical experiments we use a bi- or tri-linear interpolation model and regularized cubic approximation methods provided in FAIR to obtain the intensity values of the deformed image; see~\cite{Modersitzki2009} for implementation details and other common choices.

\emph{Continuity equation}: To solve the continuity equation~\eqref{eq:continuity} we consider the \emph{Particle-In-Cell} ({\bf PIC}) method that pushes mass along the characteristics forward in time; see, e.g.,~\cite{ChertockKurganov2006}. For a given grid point $y_0\in\R^d$, we introduce a particle with its mass given by the intensity value $\CT(y_0)$. Then, we follow the trajectory of the particle along the characteristic to its final point $y_1 := y(\bfv,y_0,0,1)$. In general, $y_1$ does not coincide with a grid point. We obtain the value of the final state in a given cell by integrating the mass of all particles whose support intersects the cell. Equivalently, we compute the final density by splitting the mass of each particle among the cells adjacent to its final location. Ideally, the particles are represented by Dirac delta functions. In practice, we consider bi- or tri-linear hat functions of a certain isotropic width $\delta>0$ as proposed in~\cite{ChertockKurganov2006}.

Using the push-forward matrix $\bF$, which has also been used in~\cite{FohringEtAl2014}, this process can be written as
\begin{equation}
\label{eq:rhoEndOMT}
u(\bfx_c,1) = \bF(y(\bfv,\bfx_c,0,1)) \CT(\bfx_c).
\end{equation}

In the following, we construct the push-forward matrix in a way that ensures mass-preservation at the discrete level for any choice of $\delta $ and $h$. For ease of presentation, we describe the procedure for the one dimensional case ($d=1$). The derivation extends to tensor meshes in higher dimensions in a straightforward way under the assumption that the basis functions are piecewise polynomials in the coordinate directions. Let us assume that for $j=1,2,\ldots,m$ particles are located at the cell-centered points $\bfx_j=(j-\hf) h$ and their respective mass is given by $\bfu_{0,j} = \CT(\bfx_j)$. The particles are advected to the points $\bfy_j = y(\bfv,\bfx_j,0,1)$. For some $\delta>0$ the particles are represented by the shifted basis functions
\begin{equation*}
b^{\delta}(x,\bfy_j) =
\begin{cases}
1 - (\bfy_j-x)/ \delta  & \text{for }   \bfy_j-\delta \leq x \leq \bfy_j,\\
1 + (x-\bfy_j)/ \delta  & \text{for }   \bfy_j          <  x \leq \bfy_j + \delta,\\
0                       & \text{else,}
\end{cases}
\end{equation*}

\noindent for each $j=1,2,\ldots,n_p$. The mass of $u(\cdot,1)$ contained in the $i$th interval  $[\bfx_i, \bfx_{i+1}]$ is given by
\begin{equation}\label{eq:ukOMT}
\bfu_i(\bfv,1)
= \sum_{j=1}^{n_p} \int_{\bfx_{i}}^{\bfx_{i+1}} \bfu_j b^{\delta}(x,\bfy_j) dx
= \sum_{j=1}^{n_p} \bfu_j \left(B^\delta(\bfx_{i+1},\bfy_j) - B^\delta(\bfx_{i},\bfy_j)\right)
= \sum_{j=1}^{n_p} \bfu_j \bfF_{ij},
\end{equation}

\noindent where $B^\delta$ denotes an anti-derivative of $b^{\delta}$ and is given by
\begin{equation*}
B^{\delta}(x,\bfy_j) =
\begin{cases}
0                                            & \text{for } x < \bfy_j - \delta, \\
x - \frac{1}{2 \delta} ( 2 \bfy_j x - x^2)   & \text{for } \bfy_j-\delta \leq x \leq \bfy_j,\\
x + \frac{1}{2 \delta} (x^2 - 2 \bfy_j x)    & \text{for } \bfy_j          <  x \leq \bfy_j + \delta,\\
1                                            & \text{for } \bfy_j + \delta < x.
\end{cases}
\end{equation*}

Repeating the process outlined in~\eqref{eq:ukOMT} for all $i=1,2,\ldots,m$ yields the discrete transformed density, which is summarized in~\eqref{eq:rhoEndOMT}. Note that our scheme is \emph{mass-preserving at the discrete level by design} since exact integration is performed. In other words: the columns in $\bfF$ sum to one regardless of the choices for $\delta$ and $h$. Also note that $\bfF$ is sparse. The level of sparsity depends on the ration between $\delta$ and $h$. If we choose $n_p=m$, $\delta = h$, and $\bfx_j = h (j+1/2)$, it is easy to see that $\bfF$ is the transpose of the linear interpolation matrix~\cite{FohringEtAl2014}. Thus, the relation between~\eqref{eq:rhoEndIR} and~\eqref{eq:rhoEndOMT} mirrors the adjoint relation between the continuity and the advection equation.

\subsection{Optimization} 
\label{sub:optimization}

Using the Lagrangian methods outlined above we parametrize the final state in terms of the velocities, which we denote by $u_1(\bfv)$. We eliminate the state equation from the variational problem~\eqref{eq:variational} and---upon discretization---obtain the finite-dimensional unconstrained problem
\begin{equation}
\label{eq:discOpti}
\min_{\bfv} \left\{J(\bfv) :=  D(u_1(\bfv), \CR) + \alpha S(\bfv) \right\},
\end{equation}

\noindent where $D$ and $S$ are discrete versions of the distance measure $\CD$ and regularizer $\CS$ in~\eqref{eq:variational}. As an example, we consider the squared $\bfL_2$-distance functional. We note, that this is a modular building block of our formulation; for other choices we refer to~\cite{Modersitzki2009}. Using a midpoint rule, the discrete distance measure---also known as the \emph{sum-of-squared-differences} ({\bf SSD})---reads
\begin{equation}
\label{eq:SSD}
D^{\rm SSD}(u_1(\bfv), \CR) = \frac{h^d}{2} {\rm res}(\bfv)^\top {\rm res}(\bfv),\quad \text{ where } \quad {\rm res}(\bfv) := u_1(\bfv) - \CR.
\end{equation}

To enable a Gauss--Newton optimization, we compute the derivative of the objective function. Using the chain rule we obtain
\begin{equation*}
d_{\bfv} J(\bfv) = d_{\bfv} u_1(\bfv) d_{u_1} D(u_1(\bfv) , \CR) + \alpha d_{\bfv} S,
\end{equation*}

\noindent where the derivative of the distance measure with respect to the final state $u_1$ is for \eqref{eq:SSD} given by
\begin{equation*}
d_{u_1} D^{\rm SSD}(u_1(\bfv) , \CR) = h^d {\rm res}(\bfv).
\end{equation*}

\noindent We again refer to~\cite{Modersitzki2009} for the derivatives for other distance measures. The derivative of the regularizer can be written as
\begin{equation*}
d_{\bfv} S(\bfv) = \Delta t h^d\  (\bfB^{\top} \bfB) \bfv.
\end{equation*}

\noindent Here, $\bfB$ is a discretization of the spatial or spatio-temporal derivative operator. Similarly, the approximated Hessian is given by
\begin{equation*}
\bfH(\bfv) \approx d_2 J(\bfv)
= d_{\bfv} u_1(\bfv) \ d_2 D(u_1(\bfv), \CR) \ d_{\bfv} u_1(\bfv)^\top
+ \alpha\ \Delta t h^d  \ \bfB^\top \bfB + \gamma \bfI_n,
\end{equation*}

\noindent where $\bfI_n \in \R^{n\times n}$ denotes the identity matrix and $\gamma>0$ is a small constant to ensure positive semi-definiteness. In our numerical experiments we use $\gamma=0.01$. The Hessian of the distance measure is given by
\begin{equation*}
d_2 D^{\rm SSD} = h^3 \bfI_n.
\end{equation*}

Next, we compute the derivative of the mapping $\bfv \mapsto u_1(\bfv)$. We first consider the advection equation. Using the chain rule to differentiate~\eqref{eq:rhoEndIR} we obtain
\begin{equation*}
d_{\bfv} u_1(\bfv) = d_{\bfv} \CT(y(\bfv,\bfx_c,1,0))
= \nabla \CT(y(\bfv, \bfx_c,1,0))\  d_\bfv y(\bfv,\bfx_c,1,0).
\end{equation*}

\noindent The first term in the product is an image gradient evaluated at the end points of the characteristic curves. It is computed by differentiating the interpolation model; see~\cite{Modersitzki2009} for details. The second term is the derivative of the endpoint of the characteristic curve with respect to $\bfv$. How we compute this derivative is explained below. For the continuity equation~\eqref{eq:rhoEndOMT} we obtain
\begin{equation*}
d_{\bfv} u_1(\bfv) = d_{\bfv} \left(\bF(y(\bfv,\bfx_c,0,1)) \CT(\bfx_c) \right)
= d_y\left( \bF(y(\bfv,\bfx_c,0,1))  \CT(\bfx_c) \right)\ d_\bfv y(\bfv,\bfx_c,0,1).
\end{equation*}

\noindent The first term can be computed by differentiating the terms in~\eqref{eq:ukOMT}, for which $\bfF_{ij} > 0$, with respect to the end points of the characteristic curves. Notice that this also implies that $d_y \left(\bF(y(\bfv,\bfx_c,0,1))\right)$ is at least as sparse as the push-forward matrix.

We now present an efficient way for computing the derivative of the end point of the characteristics with respect to the velocity field. Since we use explicit time stepping schemes the derivative can be computed recursively alongside the computation of the characteristics. For example, if we use the forward Euler method in~\eqref{eq:forwardEuler} we have $d_{\bfv}y(\bfv,\bfx_c,1,0) = 0$; we obtain
\begin{equation*}
\begin{split}
d_{\bfv} y(\bfv,\bfx_c,1,t_{k+1})
= d_{\bfv} y(\bfv,\bfx_c,1,t_k)
+ & \Delta t\ d_{\bfv} I(\bfv,y(\bfv,\bfx_c,1,t_k),t_k) + \ldots\\
\ldots+ & \Delta t\ d_y  I(\bfv,y(\bfv,\bfx_c,1,t_k),t_k) \ d_{\bfv} y(\bfv,\bfx_c,1,t_k),
\end{split}
\end{equation*}

\noindent  for all $k=0,1,\ldots, N-1$. The derivatives of the interpolation scheme, $d_\bfv I$ and $d_y I$, are computed as described in~\cite{Modersitzki2009}. Notice that we do not need $d_{\bfv} y(\bfv,\bfx,1,t_k)$ in subsequent time steps. Thus, in practice, we update it directly. It is straightforward to extend this procedure to other explicit methods, such as the RK4 scheme used in our experiments; see Algorithm~\ref{alg:rk4} for details. We emphasize that neither intermediate transformations nor intermediate state variables need to be stored to compute the derivative. Therefore, the storage requirement is essentially independent of the number of time steps $N$ used to compute the characteristics. This is different to the methods described in~\cite{BegEtAl2005,MangBiros2015IS,MangBiros2016IS,MangEtAl2016SC}, which require storing at least one time-dependent scalar field to evaluate the gradient or Hessian operator.

We use a standard inexact Gauss--Newton--Krylov method for solving the finite-dimensional optimization problem. We use the implementation and stopping conditions described in~\cite{Modersitzki2009}. As to be expected, the computationally most challenging task is the computation of the search direction. Let $\bfv^i$ denote the velocity field at the $i$th iteration. Given $\bfv^i$ we obtain the search direction $\delta \bfv$ by solving
\begin{equation}
\label{eq:solveGN}
\bfH(\bfv^i) \delta \bfv = - d_\bfv J(\bfv^i).
\end{equation}

\noindent The next iterate $\bfv^{i+1}$ is computed via $\bfv^{i+1} = \bfv^i + \mu \delta \bfv$; an Armijo linesearch is performed to determine the step size $\mu$ (see, e.g., \cite{NocedalWright2006}).

We solve the symmetric and positive definite linear system in~\eqref{eq:solveGN} via a Cholesky factorization or, for large-scale problems, via iterative methods such as the conjugate gradient ({\bf CG}) method~\cite{HestenesStiefel1952}. The convergence of iterative methods depends on the clustering of the eigenvalues of $\bfH$, which can be improved by appropriate preconditioning~\cite{Saad2003}; yielding a preconditioned CG (\textbf{PCG}) method. For the examples considered in this paper, the Hessian of the regularizer is block diagonal with $d$ blocks corresponding to a discretized second- or fourth-order differential operator. Since the Hessian of the regularizer is of higher-order as compared to the Hessian of the distance term, we exploit the structure of $\bfA = \bfB^\top \bfB$ for preconditioning. Given that the velocity is discretized on a regular mesh in space and time, $\bfA$ is a structured matrix and can be written as a sum of Kronecker products of Toeplitz-plus-Hankel matrices. Thus, its pseudo-inverse can be computed efficiently using the \emph{Discrete Cosine Transform} ({\bf DCT})~\cite{HansenNagyOLeary2006}. In addition to the preconditioners available in FAIR (such as multigrid or Jacobi) we also provide an option to use $\bfA + \gamma \bfI_n$ as preconditioner; a common choice in PDE-constrained optimization problems~\cite{MangBiros2015IS,AlexanderianEtAl2016}.

The optimization problem in~\eqref{eq:discOpti} is known to be non-convex. To limit the risk of being trapped in a local minimum, we use a multilevel strategy similar to the one described in~\cite{Modersitzki2009}. First, we solve~\eqref{eq:discOpti} with a coarse discretization for the distance, regularizer, and velocities, and then refine the solution and use it as a starting guess for the optimization problem obtained on the next level. We continue this procedure until we reach a sufficiently fine discretization level, which depends on the application at hand. In addition to improving robustness, the scheme often leads to an overall reduction of computation time.

\subsection{Implementation} 
\label{sub:implementation}

We have implemented our method in MATLAB as an extension to the 2011 version of the FAIR toolbox described in~\cite{Modersitzki2009}. This allows us to exploit all distance measures, interpolation kernels, and numerical schemes provided in FAIR. A pseudocode of the RK4 method used to compute the characteristics and the derivative of the end point with respect to the velocity field $\bfv$ is given in Algorithm~\ref{alg:rk4}. It can be seen that both the characteristics and the gradient are computed in one sweep over all time points. The characteristic and the gradient can be updated in each step. This makes the memory requirements independent of the number of time steps $N$ used to compute the characteristics; the size of $d_{\bfv} u_1(\bfv)$ is $n \times m^d$. The gradient matrix is sparse. Its columns will have non-zero entries only in rows associated with discrete velocities in close proximity to the characteristic curve; we will demonstrate this experimentally; see Fig.~\ref{fig:PCG}.

The reduced memory requirement is a significant improvement over existing Eulerian or Semi-Lagrangian methods. These require storing (or recomputing) the transported images or characteristics for each time step~\cite{BegEtAl2005,MangBiros2015IS,MangBiros2016IS,MangEtAl2016SC}. The Lagrangian method proposed here, requires only the allocation of the transformed grid, which is independent of the number of time steps for the forward or adjoint solves. Further, it is possible to adapt the time step used for computing the characteristics, e.g., depending on the complexity of the trajectory.

\section{Numerical Experiments} 
\label{sec:numerical_experiments}

In this section, we demonstrate the potential of our solver based on two- and three-dimensional synthetic and real world problems of varying complexity. We compare our prototype implementation to tailored and highly optimized state-of-the-art packages for diffeomorphic image registration. For mass-preserving registration we consider the VAMPIRE package~\cite{GigengackEtAl2012}. For large deformation diffeomorphic registration we consider the hyperelastic registration model originally described in~\cite{BurgerEtAl2013, RuthottoGreifModersitzki2016}.

\subsection{General Setup} 

As stopping criteria for the Gauss--Newton optimization, we use standard settings provided in FAIR. The maximum number of inner iterations for the PCG method is set to 50; the tolerance for the relative residual is set to $0.1$.  We use a spectral preconditioner.\footnote{Since the regularization operator corresponds to a block diagonal matrix whose $4\cdot2$ blocks are discretizations of a 2D Laplacian, its pseudo inverse can be computed efficiently using DCTs (see Sec.~\ref{sub:optimization}).} The benchmark methods employ a hyperelastic regularization model, for which effective preconditioning is more challenging; see, e.g., \cite{BurgerEtAl2013,RuthottoGreifModersitzki2016}. Here, we use a matrix-free implementation of a Jacobi-PCG solver. Problem-specific parameters, such as the number of time points to represent velocity fields, times steps in the RK4 method, the image domain, the number of multi-level steps, or the padding of the domain used to represent the velocity field, are described in the respective subsections. We perform all our experiments on a x68 compute node with 40 Intel(R) Xeon(R) CPU E5-2660 processors running at 2.60GHz with a total of 256GB of memory.

We consider $H^1$ (diffusive) and/or $H^2$ (curvature) regularization models throughout our experiments. We emphasize that, as we have already pointed out in Sec.~\ref{sub:regularization_functionals}, theoretical considerations require imposing $H^2$-regularity on $v$ in order to guarantee that $v$ gives rise to a diffeomorphic map $y$ (see, e.g.,~\cite{BegEtAl2005}). Our argument for also considering an $H^1$ regularization model is that, in practice, we can control the weight $\alpha$ by monitoring $\det \grad y$ to ensure that the discretized map $y$ is indeed a diffeomorphism. We also note that the regularization is a modular block of our formulation. If theoretical requirements are of concern, one can switch to $H^2$ regularity.


\subsection{2D C-Shape} 
\label{sub:2d_lddmm_example}

We consider the classical test case of registering a C-shaped object to a disc as initially proposed by Christensen~\cite{Christensen:1996a}. We study registration quality and performance. We compare our results to the hyperelastic registration method described in~\cite{BurgerEtAl2013,RuthottoGreifModersitzki2016}. We also study the convergence for different types of preconditioners for this problem.

\paragraph{Experimental Setup} 

The test data is taken from FAIR and consists of two binary image data with $128\times128$ pixels on the image domain $\Omega = (0,1)^2$. To build a continuously differentiable image model from the binary image data, we use the moments-regularized cubic B-spline interpolation with an experimentally tuned smoothing parameter of $\theta=0.1$; see~\cite{Modersitzki2009} for details. Since the image modality is comparable in both images, we use the SSD distance measure in~\eqref{eq:SSD} to assess image similarity.

\begin{itemize}
\item LDDMM: We model the velocity field on a padded domain $(-0.5,1.5)^2$ to reduce boundary effects. We use the diffusion regularizer with an empirically determined weight of $\alpha=400$; we set $\gamma=0$. We compare results for a stationary velocity model to those obtained for a non-stationary velocity model with $n_t=2$ time intervals. We use a three-step multilevel strategy and discretize the domain for the velocities using regular meshes with $32^2, 64^2,$ and $128^2$ cells, respectively. The characteristics are computed using an RK4 method with $N=3$ time steps.  We assess registration quality and the impact of different preconditioning techniques (no preconditioning, Jacobi preconditioning, Symmetric Gauss Seidel, and spectral) on the convergence of the PCG method used to approximately solve~\eqref{eq:solveGN}. For the convergence study, we only consider the coarsest discretization level ($32\times32$ cells). The structure of the Hessian depends on the current velocity estimate. We compare the convergence of the PCG method at the first and final Gauss--Newton iteration. We consider an $H^1$ regularization model. We report results for a stationary and a non-stationary velocity field ($n_t=2$). In each case, we aim to solve the linear system up to a relative error of $10^{-10}$ and set the maximum number of iterations to 250.
\item Hyperelastic Registration~\cite{BurgerEtAl2013}: We use the default parameters provided in FAIR to solve this problem. The values for the regularization are empirically chosen and set to $\alpha_1 = 100$ for the length $\alpha_2=0$ for the area, and $\alpha_3 = 18$ for the volume regularizer. We employ a five-step multilevel strategy, where the transformation is discretized on meshes with $8^2, 16^2, 32^2, 64^2,$ and $128^2$ cells.
\end{itemize}


\paragraph{Observations} 

For the proposed method with a stationary velocity model we require 16, 4, and 4 Gauss--Newton iterations per level with a total runtime of roughly 6 seconds. Using the non-stationary velocity model we require 25, 3, and 3 iterations per level and a runtime of about 12 seconds. For the hyperelastic registration approach 30, 17, 6, 7, and 3 iterations are performed on each resolution level. The total computational time is about 35 seconds.

\begin{figure}
\begin{center}
\includegraphics[width=.8\textwidth]{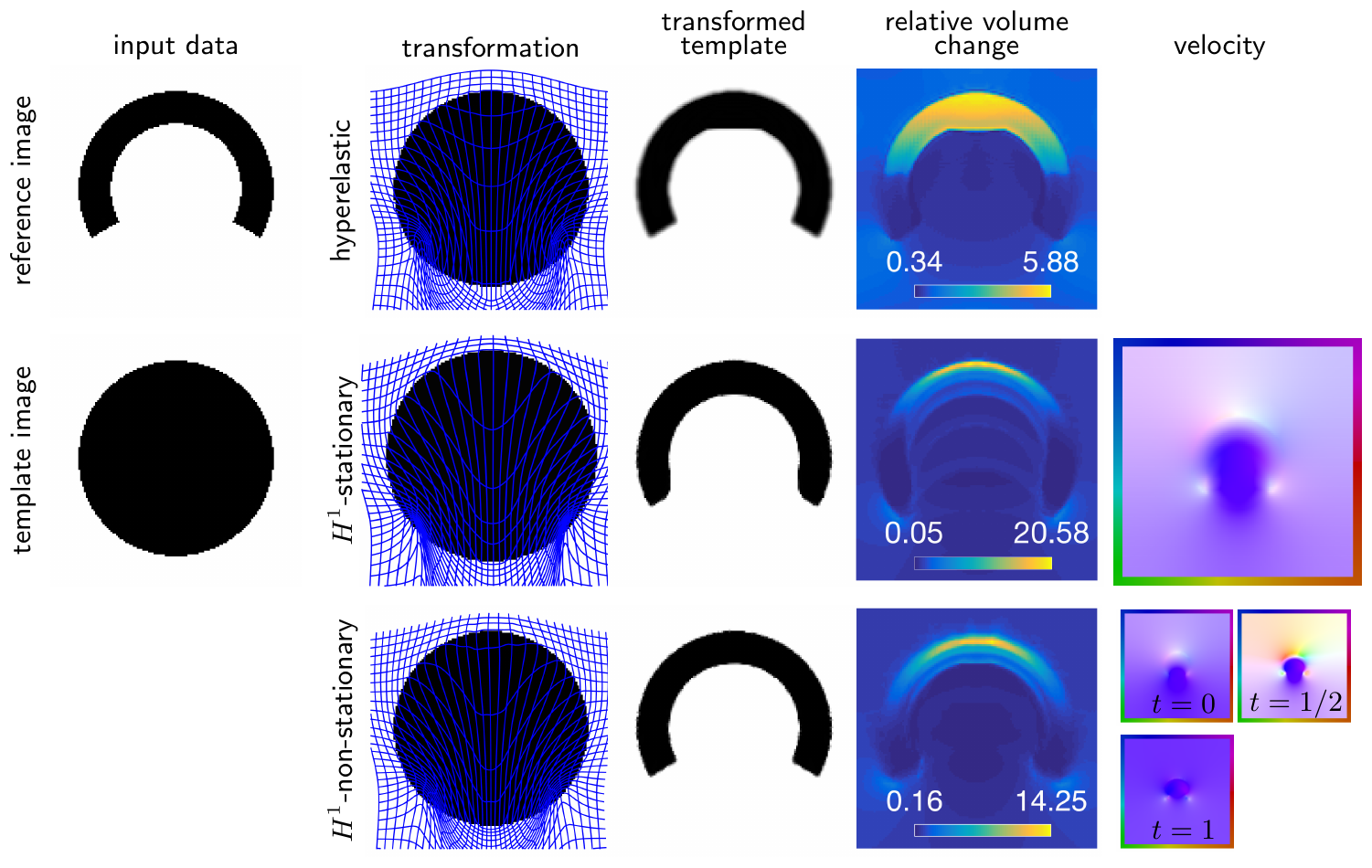}
\end{center}
\caption{2D registration results for an academic benchmark problem also considered in~\cite{Christensen:1996a}. First column visualizes test data and the remaining images visualize registration results for hyperelastic registration~\cite{BurgerEtAl2013} (first row) and the proposed method with $H^1$-regularization and stationary (second row) and non-stationary (third row) velocity models.  It can be seen that the proposed methods improves the similarity between the reference and the transformed template image without foldings of the grid. However, the ranges of the relative volume change is considerably larger. It is also evident that the non-stationary velocity model improves the registration result and comparing the estimated velocity fields (right column) shows substantial differences of the velocity estimates. }
\label{fig:CircleToC}
\end{figure}

We visualize the results in Fig.~\ref{fig:CircleToC}. As can be seen in Fig.~\ref{fig:CircleToC}, the proposed methods deliver transformed template images that are qualitatively similar to the reference image (small residual). Both methods result in diffeomorphic transformations as judged by the values of the determinant of the Jacobian. As to be expected, the range of the relative volume change is considerably larger for the proposed methods ($\det\nabla y(\bfv,\bfx,1,0) \in [0.05,20.58]$ and $\det\nabla y(\bfv,\bfx,1,0) \in [0.16,14.25]$ for the stationary and non-stationary field, respectively) as compared to the hyperelastic registration ($\det\nabla y \in [0.34,5.88]$). This is due to the fact that the hyperelastic registration model explicitly controls and penalizes volume change. Comparing the estimated stationary and non-stationary velocity fields shows that for the latter the estimate changes considerably in time. Also, it should be noted that the registration quality is slightly better for the non-stationary approach (smaller range for the Jacobians and a reduction of the distance measure of $99.48$\% vs. $97.85$\%, respectively).

\begin{figure}
\centering
\includegraphics[width=\textwidth]{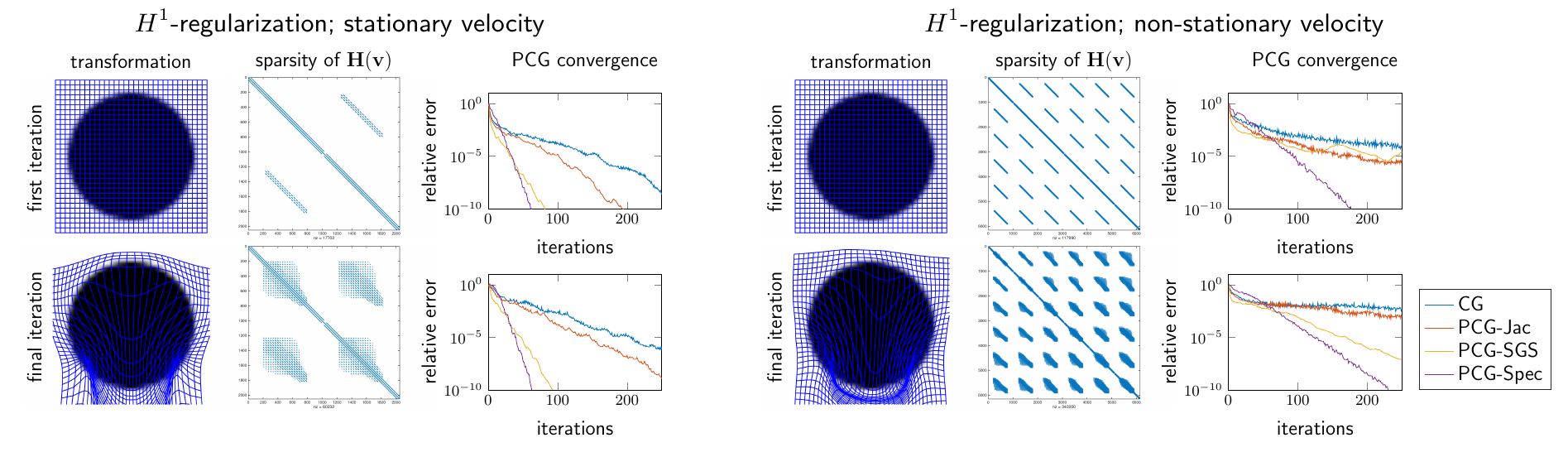}
\caption{Sparsity pattern and PCG convergence plots at first and final Gauss--Newton iteration for the 2D test problem on a coarse mesh ($m=[32,32]$). We compare the the stationary (left) and non-stationary (right) diffusion regularizer. In both cases the number of non-zero elements in the Hessian grows between the iterations since particles move farther through the domain. In all four cases  we compare the convergence of different PCG schemes (no preconditioning, Jacobi, Symmetric Gauss Seidel, and spectral preconditioning). It can be seen that the problems at the final iteration are, in this example, more difficult to solve, however, the spectral preconditioner outperforms the other choices.}
\label{fig:PCG}
\end{figure}

We show the results of the experimental evaluation of different preconditioning techniques in Fig.~\ref{fig:PCG}. The number of non-zero elements in the Hessian increases from the first to the final iteration for both regularizers. This is due to the fact that particles travel a longer distance through the domain. The performance of the preconditioner deteriorates in the final iteration for all preconditioners. We observe a similar behavior for the hyperelastic formulation (see~\cite{RuthottoGreifModersitzki2016}). We can observe that we need fewer iterations for the stationary case to reach the tolerance; we invert for fewer unknowns, which results in a smaller linear system that needs to be solved. The proposed spectral preconditioner displays the best rate of convergence amongst the considered schemes for preconditioning the Hessian; we use it for all our experiments. We note that we have performed the same study for the curvature regularization (results not included in this study). We observed a similar behavior.


\subsection{2D Mass-Preserving Registration} 
\label{sub:2d_mass_preserving_registration}

We consider an academic test problem for mass-preserving registration. We compare our method against the VAMPIRE toolbox for mass-preserving registration~\cite{GigengackEtAl2012}.

\paragraph{Experimental Setup} 

The test data is designed to mimic the contraction of a tissue containing a fixed amount of tracer. The data is obtained by subtracting two Gaussians with different standard deviations. The mass is exactly equal, but in the reference image it is concentrated in a smaller region so that the image overall appears brighter. The image domain is $(-5,5)^2$ and the full resolution is $256\times 256$. For all experiments we use a four-level multi-level strategy with resolutions $32^2$, $64^2$, $128^2$, and $256^2$, respectively. A continuous image model is built using bi-linear interpolation and the SSD distance measure is used to quantify image similarity.

\begin{itemize}
\item MP-LDDMM: As in the previous example the velocity field is modeled on a padded spatial domain $(-5.4,5.4)^2$ to reduce boundary effects. We use $n_t = 1$ for the spatial discretization of the velocity. The characteristics are approximated using $N=2$ time steps for the RK4 method. The push-forward matrices are build from bilinear basis functions whose width equals the cell size. We use the diffusion regularizer with weight $\alpha=1000$ and $\gamma=\num{1E-2}$. We compare results for a stationary and a non-stationary velocity field.
\item VAMPIRE: We use the default parameters for the hyperelastic regularizer ($\alpha_1= 10,000$ for the length-, $\alpha_2=0$ for the area, and $\alpha_3=100$ for the volume regularizer).
\end{itemize}


\paragraph{Observations} 

\begin{figure}[t]
\centering
\includegraphics[width=0.9\textwidth]
{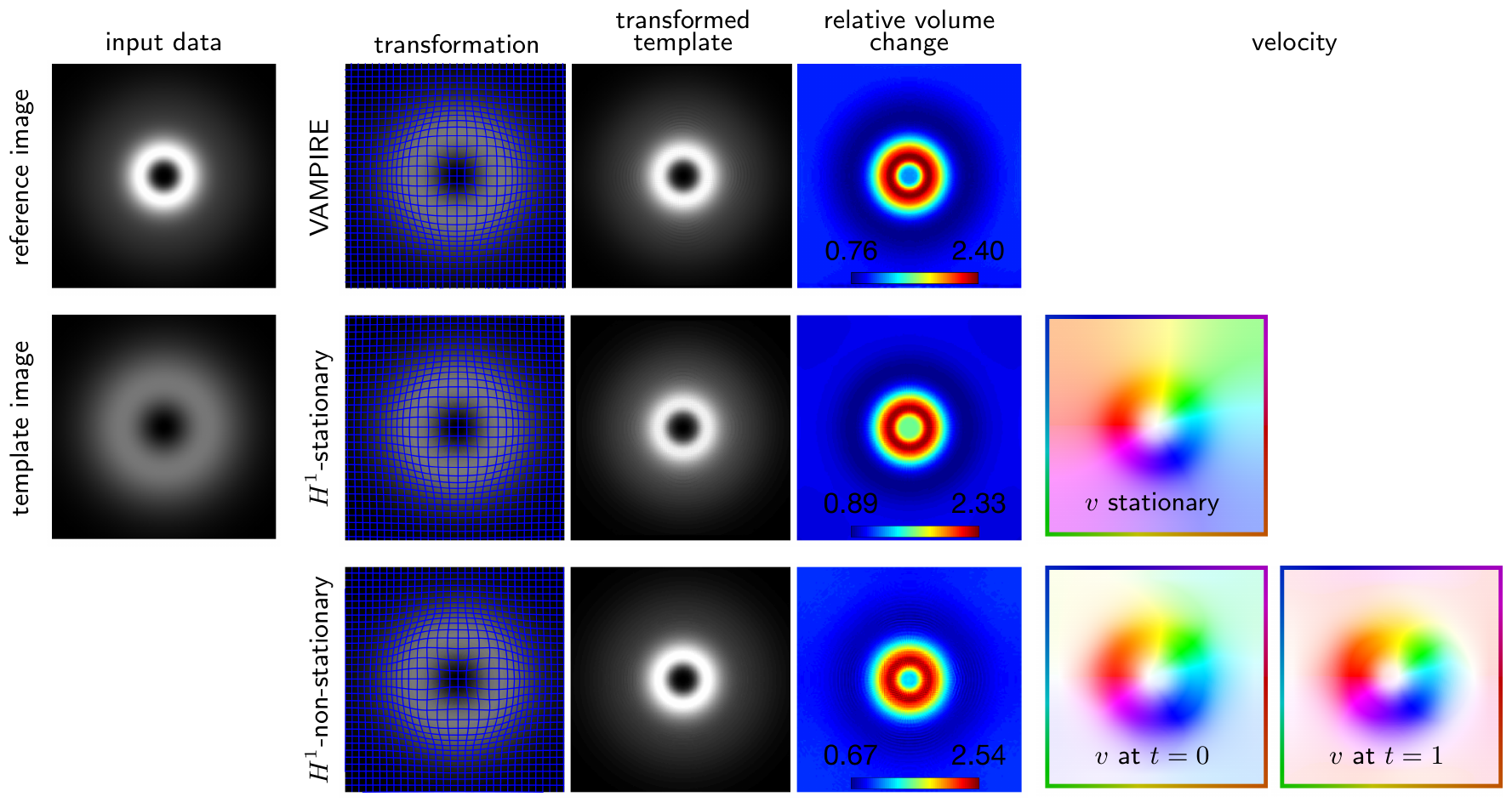}
\caption{2D mass-preserving registration for an academic test problem. The image data is generated by subtracting two Gaussian kernels with different standard deviations; the data is designed to have equal mass. The reference image (top) and the template image (bottom) are shown in the left column. We compare the VAMPIRE method (first row)~\cite{GigengackEtAl2012} to the proposed mass-preserving LDDMM with  stationary (middle row) and non-stationary velocity model (bottom row). For all three methods, we visualize the transformation, the transformed template, and the relative volume change. For the LDDMM methods we also visualize the velocity. Comparing the results in the middle and bottom row, it can be seen that the underlying transformation is rather simple; it can be well represented using a stationary velocity field.\label{fig:Gaussian}}
\end{figure}

Registration results are visualized in Fig.~\ref{fig:Gaussian}. For the MP-LDDMM using a stationary velocity model, we perform 4, 2, 1, and 1 iterations per resolution level. The total computation time is about 4 seconds. Using the non-stationary velocity model we require 5, 2, 2, and 2 iterations and require a computation time of about 8 seconds. For VAMPIRE we perform 5, 2, 2, and 1 iterations on the respective levels. The time-to-solution is approximately 12 seconds.

Both methods also yield comparable results in terms of data misfit as well as the final transformation. This is not only confirmed qualitatively by visual inspection of the transformed template image and the deformed grids, but also quantitatively: The volume change introduced by the transformation obtained using the proposed methods  ($\det\nabla y(\bfv,\bfx,1,0) \in [0.89,2.33]$ and $\det\nabla y(\bfv,\bfx,1,0) \in [0.67,2.54]$ for a stationary and a non-stationary velocity model, respectively) is comparable to the one obtained using VAMPIRE ($\det \nabla y \in [0.76,2.40]$). The largest improvement in image similarity (with respect to the SSD) is achieved for the MP-LDDMM method with a non-stationary velocity (distance reduction of $99.98$\% vs. $97.43$\%) although---in contrast to the previous experiment---it should be noted that the estimated velocities are very similar for both approaches.


\subsection{3D Cardiac PET} 
\label{sub:3d_cardiac_pet}

\begin{figure}[t]
\centering
\includegraphics[width=0.95\textwidth]
{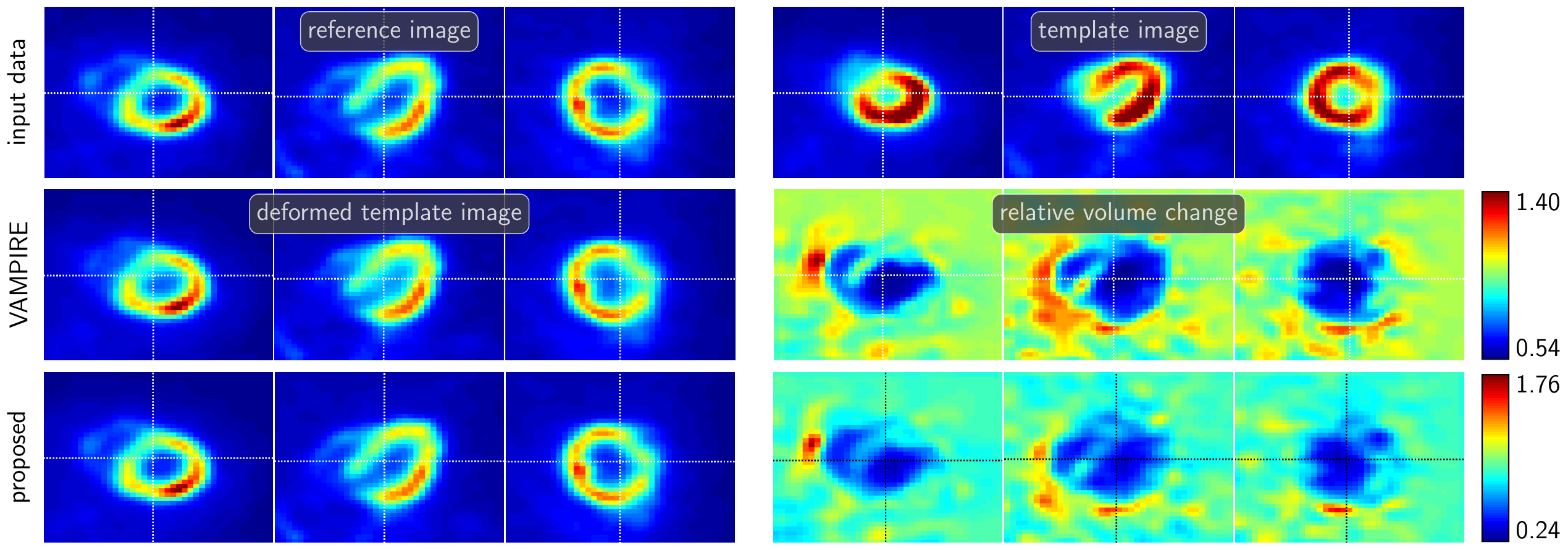}
\caption{3D mass-preserving registration of diastolic and systolic PET images of a mouse heart. We display the input data in the first row (left: reference image; right: template image; from left to right: axial, coronal and sagittal view). The deformed template images are shown in the left column (middle row: VAMPIRE; bottom row: proposed method). The relative volume change is shown in the right column (middle row: VAMPIRE; bottom row: proposed method). The color bars to the right illustrate the color coding and provide the range of the Jacobian fields.\label{fig:3Dmice}}
\end{figure}

We consider a 3D mass-preserving registration problem of registering systolic and diastolic cardiac PET data of a mouse heart. The data is provided in FAIR.\footnote{We thank the European Institute for Molecular Imaging (EIMI) and SFB 656, University of M{\"u}nster, Germany for contributing the image data.} The image domain is $\Omega = (0, 32)^3$ with a resolution of $40^3$ grid points. The results are illustrated in Fig.~\ref{fig:3Dmice}. We use a three-level multi-level strategy with resolutions $10^3$, $20^3$, and $40^3$, respectively, for all approaches. On the finest level, the number of unknowns is $\num{384000}$.

\paragraph{Experimental Setup} 
\begin{itemize}
\item MP-LDDMM: The velocity field is modeled on the same domain as the image data and the same number of cells is used for spatial discretization. We will only consider the non-stationary case, here. We use $n_t = 1$ time intervals for the velocity (which results in two discretization points for the velocity $v$). We use an RK4 method to compute the characteristics with $N = 2$ time steps. The push-forward matrix is build using tri-linear hat functions, with a width that corresponds to the voxel size of the image data. We use the diffusion regularizer with regularization weight $\alpha=\num{100}$ and $\gamma=\num{1E-2}$.
\item VAMPIRE: We use $\alpha_1 = \num{100}$ for the length regularizer, $\alpha_2=\num{10}$ for the area regularizer, and $\alpha_3=\num{100}$ for the volume regularizer. We use the same number of multi-resolution levels.
\end{itemize}


\paragraph{Observations} 
\label{par:3DmiceComparison}

For MP-LDDMM the optimization scheme performs 9, 3, and 3 iterations on the respective levels. The time-to-solution is about 36 seconds. For VAMPIRE we require 5, 4, and 3 iterations on the respective levels. The total runtime is about roughly 74 seconds. Both schemes yield qualitatively almost identical results. The residual differences between the transformed template image and the reference image is small. Overall, our current prototype implementation of a Gauss--Newton--Krylov method for LDDMM is competitive with VAMPIRE in terms of the runtime. We expect to be able to drastically reduce the runtime in near future. For the hyperelastic registration most time is spent on determining the search direction, which requires solving an ill-conditioned linear system; see also~\cite{RuthottoGreifModersitzki2016}; a reduction in runtime for this scheme is much more difficult.

\vspace{5mm}

\subsection{3D Brain Registration} 
\label{sub:3d_brain_nirep}

\paragraph{Experimental Setup} 

The data is taken from the NIREP repository~\cite{Christensen:2006a}. We consider the datasets {\tt na02} (template image) and {\tt na01} (reference image) for our experiments. The grid size for these images is $256\times300\times256$. We downsample these images to a size of $128\times150\times128$ voxels to make the problem computationally tractable for our prototype and the reference implementation. The image domain is defined to be $\Omega = (0,20) \times (0,23.4375) \times (0,20)$. We use SSD as distance measure and a multi-level strategy with 3 resolution levels ($32\times 38 \times32$, $64\times75 \times64$, and $128 \times150 \times128$). The number of unknowns is $\num{7372800}$ for the finest level.

We evaluate registration performance based on overlap measures evaluated for the label maps associated with the images. The data comes with 32 labels for gray matter regions~\cite{Christensen:2006a}. We simplify the presentation of our results by only considering the union of these labels to evaluate the performance of our method. We use the Dice coefficient as a measure for registration quality, which has an optimal value of 1. We use a nearest-neighbor interpolation model to transform the label maps with the computed $y$ to avoid any additional thresholding.

We limit the evaluation of the determinant of the Jacobian to the foreground (i.e., the brain) in the reference image. We identify this foreground by thresholding; we consider intensities with a value of 0.05 and larger as foreground. We slightly extend this mask by smoothing it with a Gaussian kernel of width $2h$. A second thresholding step defines the final brain mask used for the evaluation of the Jacobians.

\begin{itemize}
\item Proposed (LDDMM): The velocity field is modeled on a slightly larger domain than the image domain to reduce boundary effects; we choose $\Omega^v = (-1,21) \times (-1,24.4375) \times (-1,21)$. We consider stationary and non-stationary velocities $v$. We use $n_t = 1$ time intervals for the non-stationary case (which results in two discretization points for the velocity $v$). We use an RK4 method with $N = 5$ time steps to compute the characteristics . The push-forward matrix is build using tri-linear hat functions, with a width that corresponds to the voxel size of the image data. We consider the curvature ($H^2$) and the diffusive ($H^1$) regularization model. We study registration performance (data mismatch and extremal values of the Jacobians $\det\grad y$) as a function of the regularization weight $\alpha$. Once we have found the velocity $v$, we compute the transformation $y$ we use to evaluate the performance of our method using $N=20$ time steps. We experimentally found that a shift of $\gamma = 0$ and $\gamma=\num{1E-2}$ yields the optimal rate of convergence for the diffusive and the curvature regularization model, respectively. We set the tolerance for the optimization to $\text{tol}_J=\num{5E-2}$. We use a relative tolerance of \num{1E-1} for the PCG method; we limit the number of Krylov iterations to 50.
\item Hyperelastic registration: We experimentally found that regularization weight of $\alpha_1=100$ (length regularizer), $\alpha_2 = 10$ (surface regularizer), and $\alpha_3= 100$ (volume regularizer) yields high data fidelity (good mismatch) and well behaved Jacobians. We use this setting throughout our experiments. We set the tolerance for the optimization to $\text{tol}_J=1E-3$.
\end{itemize}


\begin{figure}
\centering
\includegraphics[width=\textwidth]
{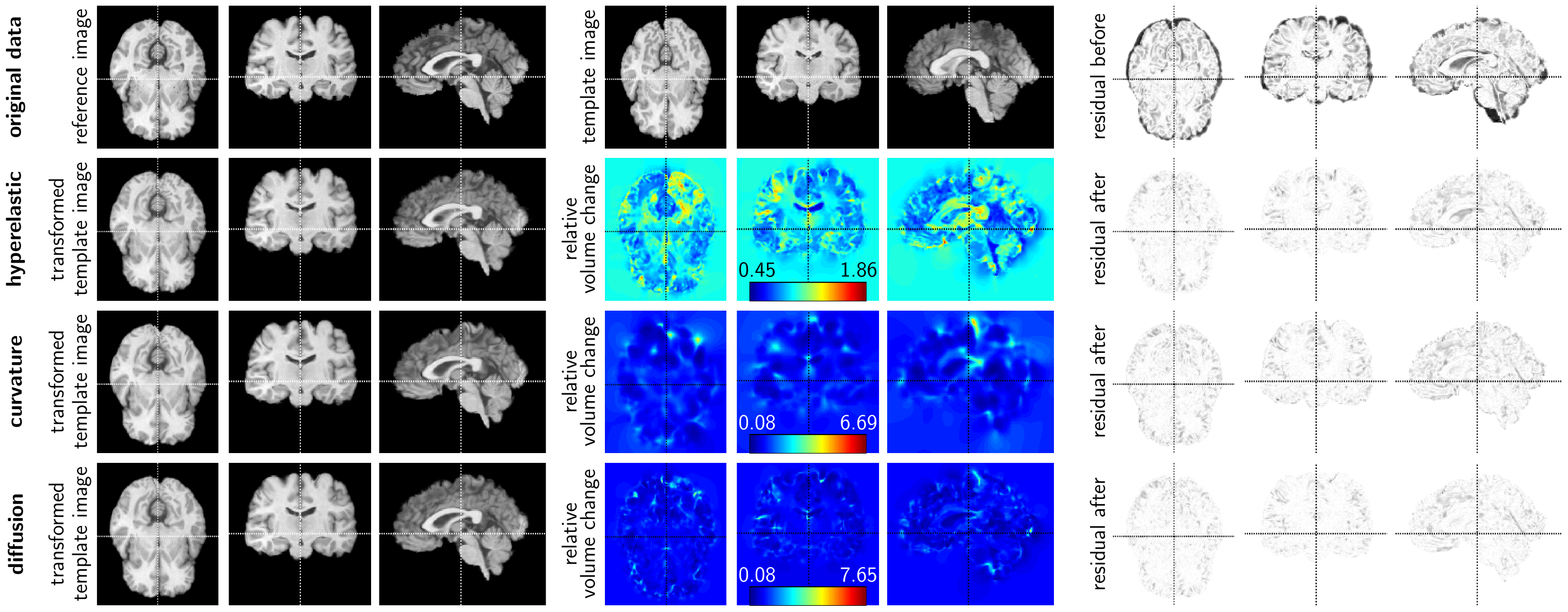}
\caption{Exemplary results for a 3D intensity-preserving registration problem based on MRI datasets of the human brain. The data is taken from the NIREP repository. We show (from left to right) an axial, coronal, and sagittal view of the reference image (dataset {\tt na01}), the template image (dataset {\tt na02}), and the residual differences between these two images in the top row. The results correspond to those reported in Table~\ref{tab:quantitative-nirep}. We report results for a map based approach with a hyperelastic regularization model (second row: run \#1 in Table~\ref{tab:quantitative-nirep}; $\alpha_1=100$ (length regularizer), $\alpha_2 = 10$ (surface regularizer), and $\alpha_3= 100$ (volume regularizer))~\cite{BurgerEtAl2013}, and for the proposed method for a non-stationary velocity field (third row: curvature regularization model; $\alpha = 10$; run \#10 in Table~\ref{tab:quantitative-nirep}; bottom row: diffusive regularization model; $\alpha = 300$; run \#13 in Table~\ref{tab:quantitative-nirep}). For each of these methods we show (from left to right) an axial, a coronal, and a sagittal view of the deformed template image, a map for the relative volume change, and the residual differences between the transformed template image and the reference image after registration. We also display the color bar and the maximal and minimal values for the maps for the relative volume change.\label{fig:3Dnirep}}
\end{figure}

\begin{table}
\caption{Registration quality. We report registration results as a function of the regularization weight $\alpha$ for the first two datasets of the NIREP repository. We compare the proposed method considering stationary and non-stationary velocity fields. We report results for the curvature ($H^2$) regularization model (theoretically required to guarantee the existence of a diffeomorphic deformation map) and a diffusive ($H^1$) regularization model. We compare the proposed method to a formulation for diffeomorphic image registration based on a hyperelastic regularization method~\cite{BurgerEtAl2013}. We use the experimentally determined regularization weights $\alpha_1=100$ (length regularizer), $\alpha_2 = 10$ (surface regularizer), and $\alpha_3= 100$ (volume regularizer). We report values (from left to right) for the Dice coefficient after registration and the extremal values for the Jacobian ($\min$ and $\max$). The initial value for the Dice coefficient for the considered datasets is \num{5.543534386E-01}.\label{tab:quantitative-nirep}}
\scriptsize\centering
\begin{tabular}{llrRrrrl}\toprule
method       & run    & $\alpha$     & dice                  & $\min(\det\nabla y)$  & $\max(\det\nabla y)$  & time (speedup)        & \\\midrule
hyperelastic & \runid & 100, 10, 100 & \num{7.933170596E-01} & \num{4.513155512E-01} & \num{1.862803169E+00} & \num{3.706054800E+03} & (\num{1.000000000})\\\midrule
\multicolumn{8}{c}{stationary LDDMM}                                                                                                                      \\\midrule
curvature    & \runid &  10          & \num{7.764123715E-01} & \num{8.876992884E-02} & \num{8.793200978E+00} & \num{1.816074592E+03} & (\num{2.040695254})\\
             & \runid &  25          & \num{7.548417772E-01} & \num{1.193601635E-01} & \num{5.766703129E+00} & \num{1.855497767E+03} & (\num{1.997337246})\\
             & \runid &  50          & \num{7.336621571E-01} & \num{1.789841204E-01} & \num{4.735466508E+00} & \num{1.447520290E+03} & (\num{2.560278309})\\
             & \runid & 100          & \num{7.141735014E-01} & \num{2.369603205E-01} & \num{3.132339125E+00} & \num{1.474501181E+03} & (\num{2.513429523})\\
diffusion    & \runid & 300          & \num{7.860163795E-01} & \num{7.789530517E-02} & \num{6.380539417E+00} & \num{1.957822580E+03} & (\num{1.892947215})\\
             & \runid & 400          & \num{7.750227852E-01} & \num{1.017359434E-01} & \num{6.264590010E+00} & \num{1.941301868E+03} & (\num{1.909056423})\\
             & \runid & 500          & \num{7.659294175E-01} & \num{1.178093780E-01} & \num{6.166939898E+00} & \num{1.948818006E+03} & (\num{1.901693636})\\\midrule
\multicolumn{8}{c}{non-stationary LDDMM}                                                                                                                  \\\midrule
curvature    & \runid &   5          & \num{7.595132911E-01} & \num{9.545388132E-02} & \num{6.357335053E+00} & \num{7.640378580E+03} & (\num{0.485061671})\\
             & \runid &  10          & \num{7.595373122E-01} & \num{8.480122115E-02} & \num{5.726130378E+00} & \num{7.498044615E+03} & (\num{0.494269505})\\
             & \runid &  25          & \num{7.467159258E-01} & \num{1.310324124E-01} & \num{5.677905399E+00} & \num{6.178465357E+03} & (\num{0.599834196})\\
             & \runid &  50          & \num{7.350425329E-01} & \num{2.123510134E-01} & \num{3.680210984E+00} & \num{5.980091501E+03} & (\num{0.619732123})\\
diffusion    & \runid & 300          & \num{8.053458765E-01} & \num{7.705591708E-02} & \num{6.475742122E+00} & \num{3.293675673E+03} & (\num{1.125203319})\\
             & \runid & 400          & \num{7.925344925E-01} & \num{1.076371450E-01} & \num{5.395472375E+00} & \num{3.342763690E+03} & (\num{1.108679866})\\
             & \runid & 500          & \num{7.672242420E-01} & \num{1.528603048E-01} & \num{6.498834778E+00} & \num{2.588212381E+03} & (\num{1.431897485})\\
\bottomrule
\end{tabular}
\end{table}

\begin{figure}
\centering
\includegraphics[width=.98\textwidth]
{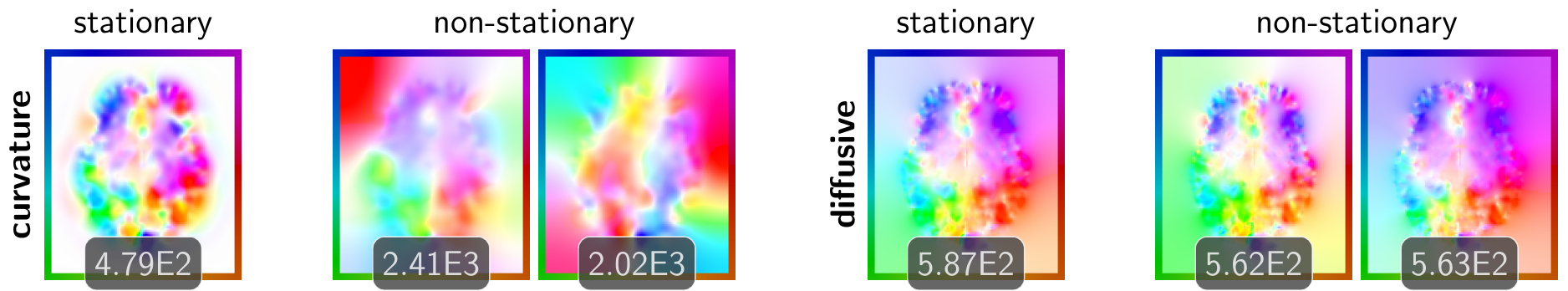}
\caption{Illustration of the obtained velocity fields for the registration of 3D brain imaging data. We show the velocities for the curvature (left;  $\alpha = 10$; runs \#2 and \#10 in Table~\ref{tab:quantitative-nirep}) and the diffusive regularization model (right; $\alpha = 300$; runs \#6 and \#13 in Table~\ref{tab:quantitative-nirep}). We report the $\ell^2$-norm of the velocity field below each individual figure.\label{fig:3Dnirep-velocities}}
\end{figure}

\paragraph{Observations} 

We show exemplary results for the registration in Fig.~\ref{fig:3Dnirep}. We report results for the quantitative evaluation in Table~\ref{tab:quantitative-nirep}. An illustration of the velocities can be found in Fig.~\ref{fig:3Dnirep-velocities}.

All methods yield high fidelity results with diffeomorphic transformations and well behaved Jacobians. We achieve the best Dice score for a diffusive regularization model for $\alpha=300$ (run \#13 in Table~\ref{tab:quantitative-nirep}) and a non-stationary velocity field. This is the only run, for which we outperform the hyperelastic approach. The results for the curvature regularization model do not vary significantly when switching from a stationary to a non-stationary formulation; we obtain similar extremal values for the Jacobians and similar Dice values. This is different for the diffusive regularization model. We obtain slightly better values for the Dice coefficient with similar extremal values for the Jacobian.

If we consider a stationary velocity field we can reduce the time-to-solution by a factor of two compared to the hyperelastic approach. These findings are consistent for both regularization approaches (runs \#2 to \#8 in Table~\ref{tab:quantitative-nirep}). If we turn to non-stationary velocity fields, our current implementation of the curvature regularization model is no longer competitive in terms of time-to-solution. For a diffusive regularization model we are, however, still slightly faster than the hyperelastic approach (runs \#13 to \#15 in Table~\ref{tab:quantitative-nirep}) despite an increase of the number of unknowns by a factor of 2 (we use $n_t=1$, which results in two discretization points for the velocity).

We need, for instance, 5, 3 , and 3 iterations for the individual levels for the stationary case and a diffusive regularization model ($\alpha=400$; run \#6 in Table~\ref{tab:quantitative-nirep}). For each iteration we require 22, 23, 24, and 24 PCG iterations (level 1), 22, 25, and 27 PCG iterations (level 2), and 27, 29, and 29 PCG iterations (level 3), respectively. The stationary case and a curvature regularization model ($\alpha = 50$; run \#4 in Table~\ref{tab:quantitative-nirep}) requires 5, 3, and 2 iterations, with 7, 11, 17, 16, and 16 PCG iterations (level 1), 20, 36, 50 PCG iterations (level 2), and 50, and 50 PCG iterations (level 3), respectively. The hyperelastic regularization approach converges after 7, 6, and 5 iterations per level.

The results in Fig.~\ref{fig:3Dnirep} suggest that all methods yield comparable residual differences after registration. However, we can, likewise to the former experiments, observe drastic differences in the Jacobians. The hyperelastic regularization allows us to better control the Jacobians (the values range from \num{4.513155512E-01} to \num{1.862803169E+00}). If this control is indeed of importance in practical applications remains to be seen. Notice, that we can either add hard constraints on the divergence of the velocity to our formulation~\cite{MangBiros2015IS, MangBiros2016IS} or constraints on $\det\grad y$ to enable such control.

The projections of the velocity fields in Fig.~\ref{fig:3Dnirep-velocities} show significant differences between the stationary and the non-stationary case for the curvature regularization model. We can also observe large differences in the appearance of the velocity fields in time for the non-stationary case. This is different for the diffusive regularization model. The stationary and non-stationary velocities do not differ significantly. The differences in time for the non-stationary case are also less pronounced. We can also observe that the energy for both components of the non-stationary velocity field is quite similar (as judged by the values for the $\ell^2$-norm reported in Fig.~\ref{fig:3Dnirep-velocities}). This is true for both regularization models. As to be expected we obtain much smoother velocities for the curvature regularization model.

\section{Summary and Conclusion} 
\label{sec:summary_and_conclusion}

In this paper, we propose efficient numerical algorithms based on Lagrangian hyperbolic PDE solvers to efficiently solve the reduced formulation of the PDE-constrained optimization problem arising in LDDMM. Our formulation can be used for classical, intensity-preserving, registration but also extends the LDDMM framework to mass-preserving registration problems. We consider an optimal control formulation and propose an efficient discretize-then-optimize approach amendable for standard Gauss--Newton methods. The key idea of our approach is to eliminate the hyperbolic PDE constraint using a Lagrangian method with an explicit time integration of the characteristics. Our formulation can handle both stationary and non-stationary velocity fields efficiently. We present economical schemes for analytically computing its derivatives. A main advantage of our method over existing solvers is that derivatives of the solution to the hyperbolic PDE with respect to the velocity field can be explicitly constructed. This leads to an overall memory requirement that is independent of the number of time steps used for solving the PDE. This is a significant advantage over most existing work, which in general require the storage (or re-computing) of spatio-temporal state and adjoint fields or the transformation.

We studied registration performance considering different synthetic and real-world problems. We made the following observations:
\begin{itemize}
\item Our results are competitive in terms of both time-to-solution and inversion quality (mismatch) compared to state-of-the-art packages for diffeomorphic image registration across a wide range of applications, which includes mass-preserving and intensity-preserving registration problems.
\item Our spectral preconditioner yields a good performance. However, the rate of convergence deteriorates when switching from stationary to non-stationary velocities. Designing a more effective preconditioner for these cases is an item of future work.
\item We could observe differences between the stationary and non-stationary formulation in terms of the reconstruction accuracy. This especially becomes apparent for the classical problem of registering a C-shaped object to a disc~\cite{Christensen:1996a}. In this example a considerable improvement can be achieved using a small number of time discretization points for the velocity. In general, increasing the number of time points enriches the space of transformations, however, it also increases the complexity of the optimization problem.
\item Since $y$ appears explicitly in our formulation, we can control $\det \nabla y$ by adjusting the regularization weight $\alpha$ (additional comments can be found below). We have considered $H^1$- and $H^2$-regularization norms. While theoretical considerations do require (more than) $H^2$-regularity on $v$ to guarantee that a diffeomorphic map $y$ exists~\cite{BegEtAl2005}, we could demonstrate our numerical scheme allows us to ensure that the final map $y$ is diffeomorphic at a discrete level, even for $H^1$-regularity. However, we note that we consider the regularization as a modular building block. If theoretical requirements are of concern, one can switch to $H^2$-regularity.
\end{itemize}

In theory, solutions to the variational optimal control problem are guaranteed to be diffeomorphic (under the assumption of sufficient regularity of $v$). However, as compared to other diffeomorphic registration approaches that control and thus guarantee invertibility of the \emph{discrete} transformation such as~\cite{HaberMod2006,BurgerEtAl2013}, it is more difficult to ensure this for discrete solutions to the optimal control problems. An inaccurate approximation of the characteristics may cause characteristics to cross and thus lead to non-diffeomorphic transformations. This problem is also inherent in other numerical implementations of LDDMM. Therefore, we recommend monitoring volume changes induced by the transformation to adapt the number of time steps and/or smoothness parameter. In our method, the end points of the characteristics correspond to a deformed grid, which can be analyzed or even regularized using techniques described in~\cite{HaberMod2006,BurgerEtAl2013}. Monitoring volume changes via Jacobian determinants can also be done in Eulerian or SL methods, in which the transformation is generally not computed~\cite{MangBiros2015IS,MangBiros2016IS}.

In this paper, we optimize over the velocity field $v$ instead of optimizing over the final transformation $y$, a strategy that has become predominantly used in many practical applications. Optimizing for the velocity allows us to use a fairly simple quadratic regularization model while still (in theory) ensuring invertibility of the resulting transformation. In the discrete setting, the grid might have foldings, depending on the regularization parameter and/or the accuracy of the time integration. This can be seen as a drawback compared to image registration methods that use invertibility constraints or nonlinear regularizers directly acting on the transformation. However, these regularizers are very challenging both in theory and in practice; see, e.g.,~\cite{RuthottoGreifModersitzki2016}. Another feature of more complicated regularization models such as, e.g., the elastic regularization proposed in~\cite{Fischler1973,Broit1981,DroskeRumpf2004,YanovskyEtAl2008,BurgerEtAl2013} is that they are motivated based on physical principles. The notion of plausibility of a transformation $y$ is for these type of regularization models not only limited to the prerequisite that $y$ is a diffeomorphism; it is based on bio-mechanical considerations. Thus, while achieving a very good similarity of the final images, the obtained transformation might not be plausible in all applications. However, a similar argument can be made for elasticity-based regularizers unless true material properties are known and incorporated into the regularization. Another approach to integrate bio-physical priors into diffeomorphic registration is to include more complicated state equations that model the bio-physics of a system under investigation; an example in the context of large deformation diffeomorphic image registration is the incorporation of incompressibility constraints~\cite{MangBiros2015IS,MangBiros2016IS}. This, likewise to more sophisticated regularization norms, introduces additional parameters, and as such makes an automated calibration of the method to unseen data more difficult.

Some limitations of the current method will be addressed in future work. First, for mass-preserving registration, the width of the particle kernels may be adjusted locally depending on the spacing of particles after transformation~\cite{ChertockKurganov2006}. Computing the distance to the closest neighbor is expensive, however, in our framework the Jacobian determinant is available and can be used to detect relative changes in the density of particles. Second, we will investigate locally adaptive time stepping schemes for computing the characteristics that account for the complexity of the velocity fields.


\section{Acknowledgements} 
\label{sec:acknowledgements}

AM is supported by the U.S. Department of Energy, Office of Science, Office of Advanced Scientific Computing Research, Applied Mathematics program under Award Numbers DE-SC0010518 and DE-SC0009286; and by NIH grant 10042242. LR is supported in part by National Science Foundation (NSF) award DMS 1522599.



\end{document}